\documentclass{article} 
\usepackage[main, final]{neurips_2026}
\usepackage[utf8]{inputenc} 
\usepackage[T1]{fontenc}    
\usepackage{hyperref}       
\usepackage{url}            
\usepackage{booktabs}       
\usepackage{amsfonts}       
\usepackage{amsmath}
\usepackage{nicefrac}       
\usepackage{microtype}      
\usepackage{graphicx}
\usepackage{natbib}
\usepackage{capt-of}

\usepackage[utf8]{inputenc}
\usepackage[table,dvipsnames,svgnames,x11names]{xcolor}
\usepackage{hyperref}
\usepackage{fancyhdr}
\usepackage{enumitem}
\usepackage{longtable}
\usepackage{booktabs}
\usepackage{tabularx}
\usepackage{array}
\usepackage[most]{tcolorbox}
\usepackage{listings}
\usepackage{xurl}
\usepackage{etoc}

\renewcommand{\arraystretch}{1.05}

\definecolor{Accent}{HTML}{1F4E79}
\definecolor{AccentSoft}{HTML}{EEF5FB}
\definecolor{SystemFill}{HTML}{EEF5FD}
\definecolor{UserFill}{HTML}{EEF9F0}
\definecolor{SchemaFill}{HTML}{FFF4E8}
\definecolor{TaskFill}{HTML}{F4EEFB}
\definecolor{AlignFill}{HTML}{ECFAF8}
\definecolor{NegativeFill}{HTML}{FDEEEE}
\definecolor{CodeFill}{HTML}{F6F7F9}
\definecolor{MutedBorder}{HTML}{C8CDD4}

\lstdefinestyle{promptstyle}{
  basicstyle=\ttfamily\small,
  breaklines=true,
  breakatwhitespace=false,
  columns=fullflexible,
  keepspaces=true,
  showstringspaces=false,
  upquote=true,
  tabsize=2,
  breakautoindent=false,
  breakindent=0pt
}

\tcbset{
  prompt/base/.style={
    enhanced,
    breakable,
    listing engine=listings,
    listing only,
    frame hidden,
    boxrule=0pt,
    arc=2mm,
    outer arc=2mm,
    left=2mm,
    right=2mm,
    top=1.3mm,
    bottom=1.2mm,
    listing options={style=promptstyle},
    borderline west={1.5mm}{0pt}{Accent},
    colback=CodeFill
  },
  prompt/system/.style={prompt/base, colback=SystemFill, borderline west={1.5mm}{0pt}{blue!70!black}},
  prompt/user/.style={prompt/base, colback=UserFill, borderline west={1.5mm}{0pt}{green!45!black}},
  prompt/schema/.style={prompt/base, colback=SchemaFill, borderline west={1.5mm}{0pt}{orange!85!black}},
  prompt/task/.style={prompt/base, colback=TaskFill, borderline west={1.5mm}{0pt}{violet!70!black}},
  prompt/align/.style={prompt/base, colback=AlignFill, borderline west={1.5mm}{0pt}{teal!70!black}},
  prompt/wrapper/.style={prompt/base, colback=AccentSoft, borderline west={1.5mm}{0pt}{Accent}},
  prompt/negative/.style={prompt/base, colback=NegativeFill, borderline west={1.5mm}{0pt}{red!70!black}},
  prompt/code/.style={prompt/base, colback=CodeFill, borderline west={1.5mm}{0pt}{black!45}},
  sourcebox/.style={
    enhanced,
    breakable,
    colback=black!2,
    colframe=MutedBorder,
    boxrule=0.45pt,
    arc=1.6mm,
    left=1.5mm,
    right=1.5mm,
    top=0.8mm,
    bottom=0.8mm
  }
}

\newtcblisting{PromptBlock}[1][]{prompt/code,#1}
\newtcolorbox{NoteBox}{enhanced, breakable, colback=yellow!10, colframe=yellow!45!black, boxrule=0.6pt, arc=2mm, left=2mm, right=2mm, top=1.2mm, bottom=1.2mm}

\title{Audible World Models: Spatially Aware Sound Generation for 3D Worlds}

\author{%
  \textbf{Duowen Chen} \\
  Georgia Institute of Technology \\
  \texttt{dchen322@gatech.edu}
  \And
  \textbf{Jinjin He} \\
  Georgia Institute of Technology \\
  \texttt{jhe433@gatech.edu}
  \AND
  \textbf{Gouthaman KV} \\
  Dolby Laboratories \\
  \texttt{gkv@dolby.com}
  \And
  \textbf{Sandeep Bangalore Venkatesh} \\
  Dolby Laboratories \\
  \texttt{sbang@dolby.com}
  \AND
  \textbf{Bo Zhu} \\
  Georgia Institute of Technology \\
  \texttt{bo.zhu@gatech.edu}
}

\begin{document}

\maketitle

\begin{abstract}
Text- and image-conditioned world generators can create visually rich 3D environments, yet these worlds often remain silent or rely on soundtracks synthesized solely from text or rendered video. Although such audio can convey what should be heard, it lacks an explicit representation of where sound sources are located and how their perceived sound should vary with listener movement. We introduce \emph{Audible World Models}, a training-free framework that incorporates sound into the generated world state. Starting from a text prompt, our system constructs a panoramic 3D proxy, separates it into semantic layers, identifies sound-producing foreground objects and ambient background regions, and synthesizes dry audio for each sound label. It then anchors these sources to reconstructed geometry and renders listener-dependent spatial audio using geometric acoustic propagation. By explicitly linking semantics, geometry, and sound propagation, the framework maintains persistent source locations while adapting the rendered audio to changes in listener viewpoint and motion. Experiments across 80 generated scenes demonstrate substantial gains in spatial consistency over text-, video-, and panorama-conditioned baselines, while preserving competitive semantic alignment. VLM-based assessments and human evaluations further indicate that our soundtracks are preferred for their audio–visual consistency, spatial plausibility, and motion-dependent behavior.
\end{abstract}

\section{Introduction}
Generative 3D world models are rapidly turning lightweight inputs such as text prompts or single images into explorable environments. Recent systems \cite{poole2022dreamfusion,hollein2023text2room} connect panoramic generation \cite{yu2024wonderjourney,yu2025wonderworld,liang2025wonderland}, depth-based lifting \cite{chung2023luciddreamer}, scene completion \cite{zhang2024scenedreamer3d}, and long-horizon navigation \cite{li2025wonderplay,hunyuanworld2025,sun2025worldplay,dominici2026dreamanywhere} to produce visually compelling worlds with increasing geometric consistency and user control. Yet the generated experience is usually incomplete: the world can be seen and explored, but it is silent.

Adding sound to generated worlds requires more than plausible soundtrack synthesis. Modern text-to-audio \cite{kreuk2022audiogen,liu2023audioldm,copet2023musicgen,vyas2023audiobox,evans2024stableaudioopen} and video-to-audio models \cite{luo2023diff,zhang2024foleycrafter,cheng2024mmaudio,shan2025hunyuanvideofoley} can synthesize high-quality environmental sounds, ambience, music, and Foley effects. Their conditioning is usually a prompt or short rendered video window, which gives only weak control over source identity, 3D placement, occlusion, reflection, and left--right changes under motion. Thus, a generated street may contain plausible traffic and voices while the waveform lacks a road-side source, a stable crowd region, or a geometric reason for changing as the listener turns. This gap becomes especially visible on long trajectories, camera rotations, and revisited viewpoints.

We present a world-state representation for generated audio. An audible world stores semantic source labels, dry audio assets, 3D placements, source attributes, scene geometry, and acoustic parameters before rendering. This mirrors the structural ingredients that make visual worlds explorable: layers, geometry, and motion. Classical geometric sound simulation \cite{schissler2011gsound} can render distance attenuation, occlusion, diffraction, and reverberation \cite{schissler2014highdiff}, while assuming pre-authored meshes, sources, and acoustic parameters. Generated worlds already provide approximate geometry and visual semantics, and audio generators provide sound content. Our framework connects these pieces so the same world can be heard from multiple trajectories with persistent source behavior. This turns audio into a reusable property of the generated world state.

We implement this representation with an agentic, training-free pipeline. Starting from a generated panorama, VLM agents identify audible foreground objects and ambient background regions, turn them into label-conditioned sound descriptions, synthesize dry clips, anchor sources to reconstructed 3D geometry, and render binaural audio with a geometric acoustic simulator along the listener trajectory. The agents act as semantic coordinators between visual generation, audio generation, and acoustic rendering. The result is a modular system whose components can be replaced independently and whose source-level outputs can be tested against known listener--source geometry.

We summarize our contributions as:
\begin{itemize}
  \item We formulate \emph{audible world modeling} as spatial audio generation from an explicit world state containing geometry, semantic sound labels, dry audio assets, source placements, source attributes, and acoustic parameters.
  \item We present a training-free agentic pipeline that discovers audible scene elements, synthesizes label-conditioned audio, grounds sources on reconstructed geometry, and renders listener-dependent binaural sound through geometric propagation.
  \item We evaluate geometry-referenced direction recovery for isolated sources and final mixtures, alongside semantic alignment and perceptual preference. Our method achieves $6.91^\circ$ isolated-source DoA MAE and a 67.0-percentage-point human Rank-1 preference gain over the strongest compared baseline.
\end{itemize}

\section{Related Work}
\paragraph{Text to Audio generation.}
Text-to-audio models based on autoregressive, diffusion, and flow-matching architectures \cite{chen2017deep,chen2025video} have greatly improved the quality and controllability of generated sound. AudioGen \cite{kreuk2022audiogen}, MusicGen \cite{copet2023musicgen}, AudioLDM \cite{liu2023audioldm}, TANGO/TANGO2 \cite{ghosal2023tango,majumder2024tango2}, AudioBox \cite{vyas2023audiobox}, Stable Audio 1.0 \cite{evans2024stableaudioopen}, and TangoFlux \cite{hung2024tangoflux} provide strong backbones for synthesizing sound effects, ambience, music, and general audio from natural-language descriptions. These models are useful for our setting because they can produce diverse dry audio assets; however, they do not by themselves determine persistent 3D source placement or listener-dependent acoustic effects.

\paragraph{Audio for visual worlds.}
A growing body of work conditions audio on images, videos, panoramic observations, or neural scene representations. Earlier methods infer binaural \cite{gao2018visualsound,garg2023visuallyguided} or ambisonic cues \cite{morgado2018spatialaudio360,li2018sceneaware360} from visual observations, while novel-view acoustic synthesis \cite{chen2023nvas} and neural scene methods \cite{liang2023avnerf,bhosale2024avgs,chen2024avcloud,chen2025soundvista} render audio from new listener poses using learned 3D representations. Recent systems further target video-to-spatial-audio \cite{kimvisage,dagli2024see2sound,liu2025omniaudiogeneratingspatialaudio}, panorama-conditioned audio \cite{heydari2024immersediffusion}, and 360-degree sound generation \cite{sun2024bothearswideopen,xie2025sonic4d}. Video-to-audio \cite{luo2023diff,cheng2024mmaudio} and neural Foley systems \cite{zhang2024foleycrafter,shan2025hunyuanvideofoley} provide strong semantic and temporal synchronization, but most remain conditioned on rendered observations rather than a persistent world representation. In contrast, our method explicitly stores sound sources in 3D and propagates them through the generated scene.


\paragraph{Spatial acoustics simulation.}
Physically based sound propagation has long studied how geometry, materials, and listener motion shape the acoustic signal. Interactive geometric systems such as GSound \cite{schissler2011gsound} render specular reflection, diffuse reflection, diffraction \cite{schissler2014highdiff,cao2016bst}, and multi-source propagation in dynamic scenes \cite{schissler2017multisource}. Simulation frameworks \cite{chen2020soundspaces,chen2022soundspaces2} and neural acoustic-field methods \cite{su2022inras,luo2022naf,liang2023nacf,lan2024acousticvolume} have also enabled embodied audio rendering, viewpoint interpolation \cite{wang2024hearinganything,liu2025hearinganywhere}, and learned impulse-response prediction \cite{jin2025differentiableroom}. These methods provide the physical grounding needed for spatial consistency, but they typically require manually specified scenes and sources. Our contribution is to connect this physical rendering machinery to generative world and audio models.


\begin{figure*}
    \centering
    \includegraphics[width=0.99\linewidth]{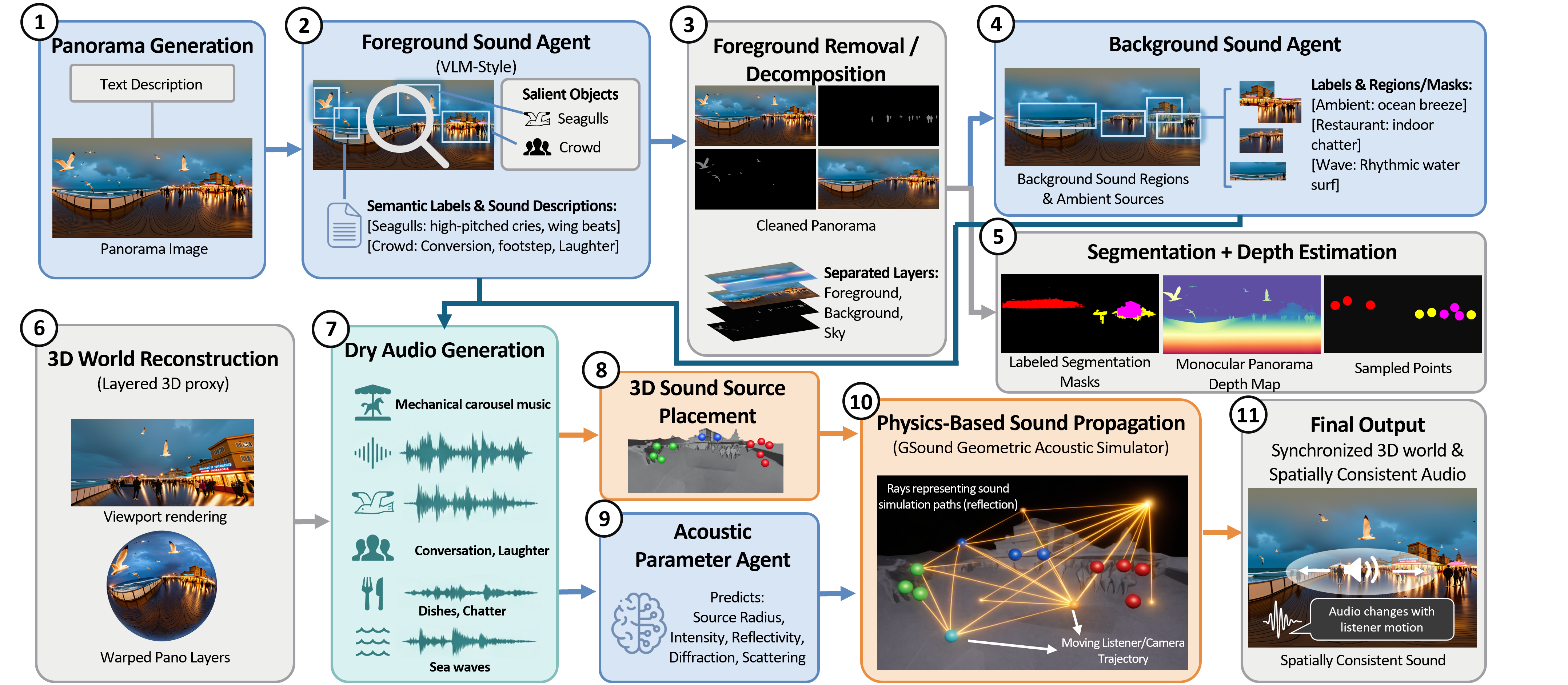}
    \caption{
    Overview of the proposed Audible World Models pipeline. The numbered blocks correspond to the stages described in Section~\ref{sec:method}. Stages 1--6 construct the sound-aware world proxy from a text prompt: a panorama, foreground audible objects, cleaned background regions, segmentation masks, depth, and layered 3D geometry. Stages 7--8 convert semantic sound labels into dry source assets and ground them as persistent 3D sources. Stages 9--11 predict acoustic parameters, propagate each source through the reconstructed geometry with GSound, and output synchronized spatial audio for the moving listener.
    }
    \label{fig:pipeline}
\end{figure*}

\section{From Visual Worlds to Audible Worlds}
\label{sec:method}
Given a text prompt \(c\) and a listener trajectory \(\tau(t)\), our goal is to generate an audio signal \(y(t)\) that is both semantically appropriate for the scene and spatially consistent with listener motion. The method is built around three design requirements. First, the sound representation should be \emph{compositional}: scene-level ambience, a river, a crowd, and a vehicle are represented as distinct components. Second, it should be \emph{persistent}: once a component is placed in the generated world, it keeps the same world-space identity as the listener moves. Third, it should be \emph{renderable}: the final binaural signal is produced from source positions, listener motion, and scene geometry.

We represent an audible world as
\begin{equation}
    \mathcal{W}_a =
    \left(
    \mathcal{G},
    \left\{(\ell_i, d_i, a_i, \mathbf{x}_i, \rho_i)\right\}_{i=1}^M,
    \Theta
    \right).
\end{equation}
The numbered pipeline in Figure~\ref{fig:pipeline} instantiates this representation. Stages 1--6 construct \(\mathcal{G}\), the semantic labels \(\ell_i\), the descriptions \(d_i\), and their visual support from the panorama. Stages 7--8 generate the dry clips \(a_i\) and convert masks, depth, and reconstructed surfaces into source placements \(\mathbf{x}_i\) and source attributes \(\rho_i\). Stages 9--11 estimate the acoustic parameters \(\Theta\) and render the final listener-dependent waveform \(y(t)\) along trajectory \(\tau(t)\).  This representation is the main distinction of our formulation: a generated world becomes a reusable sound field whose sources can be rendered from different trajectories and evaluated against known listener--source geometry. 


\subsection{Constructing a sound-aware world proxy}\label{sec:stage1}
The first part of the pipeline builds the visual and semantic support for the audible world state, as shown in stages 1--6 of Figure~\ref{fig:pipeline}.
We first generate a panoramic world proxy from the input prompt using the panoramic synthesis module of HunyuanWorld \cite{hunyuanworld2025}. The panorama gives a global observation of the generated scene before a particular perspective trajectory is rendered. This is useful for sound because many audible objects may be outside the current camera view but still affect what the listener hears. A video-window method can miss such sources until they enter view, whereas a panorama-level parse can place them in the world from the beginning.

The proxy is decomposed into foreground, background, and sky layers. The foreground pass identifies salient localized objects that may emit sound, such as vehicles, animals, crowds, machinery, musical performers, or water features. The background pass operates on a foreground-removed panorama and identifies broader ambient regions, such as forests, streets, markets, rivers, coastlines, or indoor rooms. Separating these two cases is important because they should be spatialized differently. A barking dog or passing car should remain attached to a compact location, while wind through trees or a crowd murmur may be better modeled as a distributed field over a larger region.

For each audible label \(\ell_i\), the VLM produces a concise sound description \(d_i\) and a locality attribute indicating whether the source should be modeled as localized or diffuse. The same semantic labels are used to obtain segmentation masks \cite{ravi2024sam}, which are lifted into the layered 3D representation together with monocular depth \cite{bhat2023zoedepth} and world-sheet reconstruction \cite{hu2021worldsheet,hunyuanworld2025}. The output of this stage is not an audio clip, but a structured acoustic inventory: a set of named scene entities, each linked to a visual region and a proposed sound behavior. Low-level prompt templates (Appendix~\ref{sec:foreground_prompt_app}), inpainting details and mask-processing rules (Appendix~\ref{sec:layer_mask_app}) are left to the appendix.

\subsection{Synthesizing and grounding sound sources}\label{sec:stage2}
The next part of the pipeline turns the acoustic inventory into source assets and persistent source locations, as shown in stages 7--8 of Figure~\ref{fig:pipeline}.
The parsed labels give the system a structured inventory of what should be heard. We synthesize one dry audio clip for each foreground or background sound description, plus an optional global ambience track. In the current implementation we use Stable Audio 1.0 as the text-to-audio backend \cite{evans2024stableaudioopen}, but the framework only requires a model that maps textual descriptions to audio clips. Because the audio clips are generated independently before spatial rendering, they can be treated as source assets analogous to visual objects in a 3D scene.

The key step is to attach each generated sound to the world. Localized foreground sounds are sampled on their corresponding object meshes or world sheets. Background sounds are sampled on semantic submeshes associated with their masks, and diffuse ambience is instantiated as multiple distributed sources over the relevant background support. This converts label-conditioned sound generation into a persistent source field: the same dog, fountain, crowd, or river remains at the same position as the listener moves.

This grounding step also resolves a common ambiguity in text- or video-conditioned audio generation. A prompt such as ``a lively street with traffic and people'' may imply traffic, voices, footsteps, and distant city ambience, but it does not specify which side of the listener each sound should occupy. In our representation, this ambiguity is resolved by the visual world itself: traffic is placed on the reconstructed road region, voices are placed near crowd or pedestrian regions, and diffuse city ambience is distributed across background support. Details for surface sampling, normal offsets, jitter, and source-count normalization are left to the Appendix~\ref{sec:source_placement_app}.

\subsection{Acoustic parameterization and rendering}\label{sec:stage3}
The final part of the pipeline converts grounded source assets into a synchronized spatial soundtrack, as shown in stages 9--11 of Figure~\ref{fig:pipeline}.
A geometric acoustic renderer requires more than source positions. It also needs scene-level and source-level parameters controlling reflectivity, scattering, source radius, volume, and propagation behavior. We therefore use an acoustic parameter agent that reads the scene category, audible labels, approximate source layout, and source counts, then predicts conservative acoustic settings. These estimates are clamped to stable ranges before rendering (details are provided in Appendix~\ref{sec:acoustic_params_app}). This gives the system different behavior without training a new audio model.

The agentic parameterization is intentionally lightweight. It does not attempt to recover exact material properties from the generated image; instead, it chooses plausible propagation settings that preserve the qualitative acoustic character of the scene. For example, hard urban surfaces should generally permit stronger reflections than open vegetation. This level of acoustic control is sufficient for our goal: producing spatial audio whose variation with listener motion is consistent and perceptually plausible.

Finally, we render the soundtrack using the GSound C++ API \cite{schissler2011gsound,schissler2014highdiff}. The reconstructed geometry is used as the propagation scene, the sampled sources provide dry audio and source metadata, and the camera path supplies the moving listener trajectory. The output is a binaural, listener-dependent audio signal \(y(t)\) synchronized with the generated world. Because the sound field is tied to geometry rather than only to a rendered video window, spatial effects can remain consistent over longer trajectories and revisited viewpoints. The same representation also supports diagnostic evaluation: each source can be rendered in isolation, letting us measure whether its binaural cues follow its world-space position.
\begin{table*}[t]
  \centering
  \scriptsize
  \setlength{\tabcolsep}{3.5pt}
  \renewcommand{\arraystretch}{1.12}

  \begin{minipage}[t]{0.55\textwidth}
    \centering
    \caption{Semantic evaluation on 80 generated scenes. Higher is better for all metrics.}
    \label{tab:semantic_eval}
    \resizebox{\linewidth}{!}{
    \begin{tabular}{lcccc}
      \toprule
      \textbf{Method} &
      \textbf{CLAP} $\uparrow$ &
      \textbf{IB-Text} $\uparrow$ &
      \textbf{IB-Image} $\uparrow$ &
      \textbf{Caption} $\uparrow$ \\
      \midrule
      See-2-Sound & 0.1032 & 0.0233 & 0.0245 & 0.0989\\
      OmniAudio & 0.0706 & 0.1224 & 0.0875 & 0.1055 \\
      Stable Audio 1.0 & \textbf{0.3366} & 0.1813 & 0.2477 & \underline{0.2552} \\
      MMAudio & 0.3049 & \underline{0.1953} & \textbf{0.2632} & 0.2519 \\
      Ours & \underline{0.3172} & \textbf{0.2199} & \underline{0.2618} & \textbf{0.2760} \\
      \bottomrule
    \end{tabular}
    }
  \end{minipage}
  \hfill
  \begin{minipage}[t]{0.35\textwidth}
    \centering
    \caption{DoA on final mixtures. Matched MAE is in degrees;
    $F_1$ accounts for missed and extra directions.}
    \label{tab:mixture_doa}
    \resizebox{\linewidth}{!}{
    \begin{tabular}{@{}lrrr@{}}
      \toprule
      Method & MAE $\downarrow$ & $F_1@15^\circ$ $\uparrow$
             & $F_1@45^\circ$ $\uparrow$ \\
      \midrule
      See-2-Sound & 45.6 & 0.202 & \textbf{0.526} \\
      OmniAudio & 54.35 & 0.089 & 0.265 \\
      Ours & \textbf{17.44} & \textbf{0.297} & 0.424 \\
      \bottomrule
    \end{tabular}}
  \end{minipage}
\end{table*}

\begin{table}[t]
\centering
\small
\setlength{\tabcolsep}{5pt}
\caption{Automated VLM-based evaluation using uniformly truncated
5\,s clips, including all five methods. MOS values are model-assigned
ratings, not human ratings. Higher is better except for mean rank.}
\label{tab:vlm_eval_5s}
\begin{tabular}{@{}lrrrrr@{}}
\toprule
Method & Mean MOS $\uparrow$ & Mean rank $\downarrow$
       & Rank-1 $\uparrow$ & Top-2 $\uparrow$ & Borda $\uparrow$ \\
\midrule
Stable Audio 1.0 & 2.98 & 2.59 & 21.2\% & 48.5\% & 3.41 \\
MMAudio & 2.95 & 2.68 & 24.2\% & 47.0\% & 3.32 \\
See-2-Sound & 2.93 & 3.33 & 7.6\% & 24.2\% & 2.67 \\
OmniAudio & 2.52 & 4.38 & 1.5\% & 9.1\% & 1.62 \\
Ours & \textbf{3.01} & \textbf{2.02} & \textbf{45.5\%}
     & \textbf{71.2\%} & \textbf{3.98} \\
\bottomrule
\end{tabular}
\end{table}

\begin{figure*}[t]
    \centering
    \includegraphics[width=0.99\linewidth]{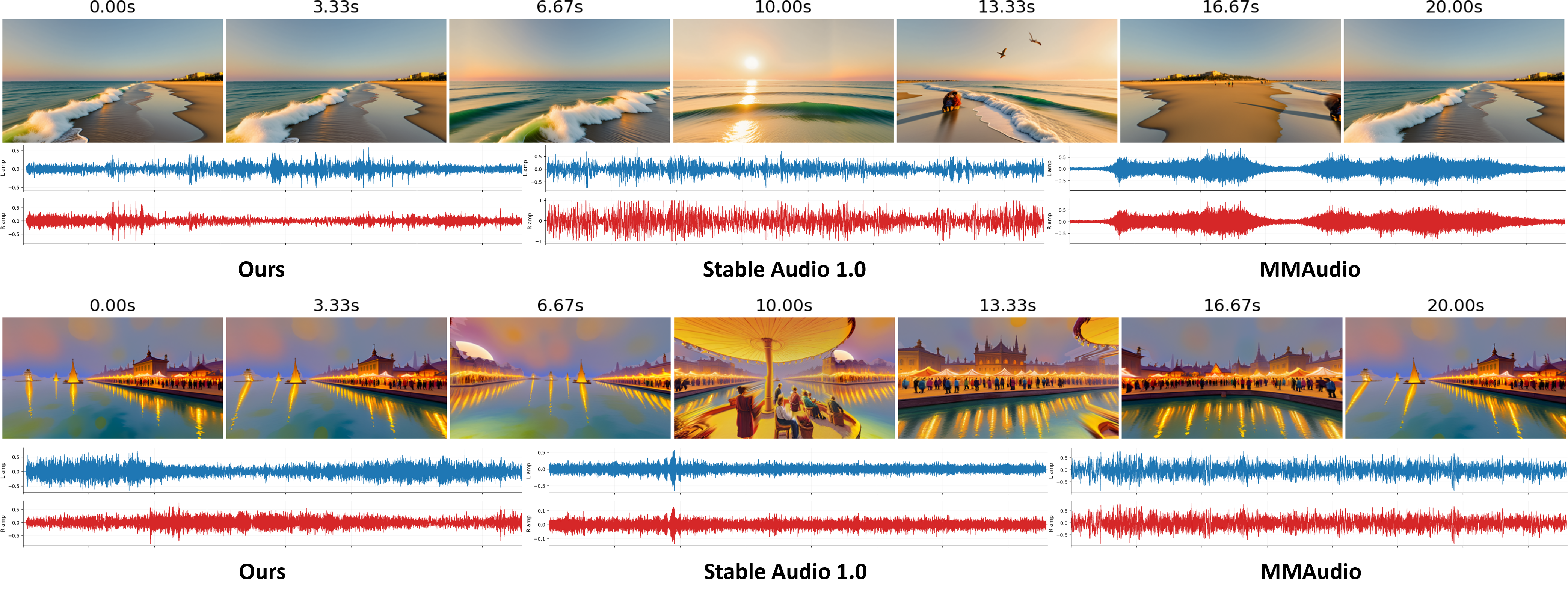}
    \caption{
    Synchronized 20-second audiovisual examples. Compared with scene-level text-to-audio and video-to-audio baselines, our method produces clearer motion-dependent changes because each sound remains tied to a persistent 3D source location.
    }
    \label{fig:sound_20s}
\end{figure*}

\begin{figure*}[t]
    \centering
    \includegraphics[width=0.99\linewidth]{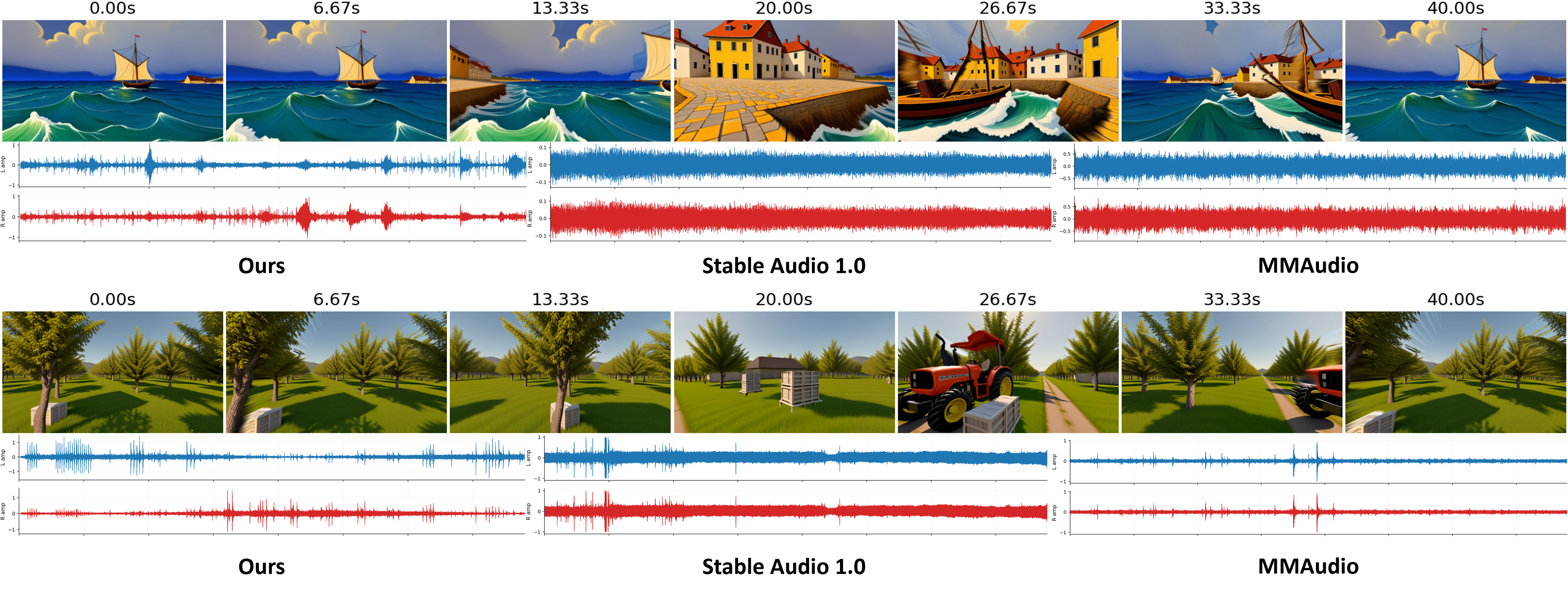}
    \caption{
    Synchronized 40-second audiovisual examples. The spatial behavior remains stable over longer trajectories, supporting the claim that explicit source placement and acoustic propagation help maintain long-horizon consistency.
    }
    \label{fig:sound_40s}
\end{figure*}

\section{Evaluation and Results}
We evaluate whether the generated soundtrack satisfies two requirements: it should contain sounds that match the scene semantics, and those sounds should change consistently as the listener moves. Our primary semantic benchmark contains 80 generated scenes, with additional controlled evaluations reported separately. Our method is training-free: no model weights are updated. The full offline pipeline is run on a single NVIDIA H200 GPU and requires up to 48\,GB of GPU memory; GSound rendering is CPU-based. End-to-end timing is reported in Appendix~\ref{sec:time_breakdown}.

We compare against a strong text-to-audio baseline, Stable Audio 1.0 \cite{evans2024stableaudioopen}; a representative video-to-audio baseline, MMAudio \cite{cheng2024mmaudio}; and panorama- or scene-conditioned spatial audio baselines, SEE-2-SOUND \cite{dagli2024see2sound} and OmniAudio \cite{liu2025omniaudiogeneratingspatialaudio}. The human study compares Stable Audio 1.0, MMAudio, and our method over 20--40\,s trajectories. The released spatial-baseline configurations used here produce shorter 5--10\,s outputs, so we evaluate all five methods separately using duration-matched 5\,s clips. Mixing output durations would confound the blinded long-horizon comparison.

\subsection{Semantic alignment}
\label{sec:semantic_definition}

Semantic evaluation asks whether the generated audio is appropriate for the target world. Stable Audio 1.0 receives the refined scene prompt, video-to-audio baselines receive a rendered trajectory video, and panorama- or scene-conditioned baselines receive the generated panorama. To isolate semantic content from spatialization, we compute semantic metrics on mono mixdowns of all outputs.

We report three complementary scores. CLAP \cite{wu2023large} measures audio--text alignment between the scene prompt and generated audio using the LAION-CLAP checkpoint adopted in prior text-to-audio evaluation \cite{liu2023audioldm,evans2024stableaudioopen,cheng2024mmaudio}. ImageBind measures cross-modal consistency with both the prompt and four rendered perspective views covering the scene \cite{girdhar2023imagebind}. The caption score first captions the generated audio using an audio captioning model \cite{ghosh2025audio} and then compares the caption with the scene prompt in a shared text-embedding space. We provide detailed metric definitions in Appendix~\ref{sec:semantic_metric_app}.

Table~\ref{tab:semantic_eval} shows that explicit source decomposition preserves strong semantic alignment. Stable Audio 1.0 obtains the highest CLAP score at 0.3366, while our method reaches 0.3172, or 94.2\% of the best score. Our method achieves the best ImageBind text score, improving from 0.1953 to 0.2199 over the strongest baseline, and the best caption score, improving from 0.2552 to 0.2760. It also reaches 0.2618 on ImageBind image alignment, within 0.5\% of the best score. These results indicate that source-level structure adds spatial grounding while keeping the audio semantically aligned with the generated world.

We additionally report scene-bootstrap 95\% confidence intervals for the
semantic metrics over the original 80 scenes; for example, the CLAP score
is 0.317 with interval $[0.286,0.348]$. Full intervals are provided in
Appendix~\ref{sec:semantic_ci_app}.

\subsection{Spatial consistency}
\label{sec:spatial_metric}
We evaluate directional consistency against source azimuths computed from
the constructed world and the known listener trajectory. This tests
whether directions are recoverable from the rendered sound, rather than
whether an ILD curve can be fitted to listener--source geometry.

\paragraph{Isolated-source direction recovery.}
We estimate full-circle horizontal direction of arrival (DoA) from
isolated binaural tracks using a calibrated combination of multi-band
ILD, interaural time difference, and a weak pinna-spectral cue for
front--back disambiguation. Our method achieves $6.91^\circ$ mean absolute
error (95\% CI: $3.51$--$12.04^\circ$), $2.96^\circ$ median error, and
96.1\% accuracy within $15^\circ$. Calibration and full intervals are
reported in Appendix~\ref{sec:doa_app}.

\paragraph{Direction recovery from mixtures.}
We additionally estimate directions directly from the final mixed
soundtracks and compare them with active reference sources using
one-to-one Hungarian matching with circular angular distance.
Table~\ref{tab:mixture_doa} reports matched angular error and $F_1$ at
$15^\circ$ and $45^\circ$ tolerances. Our method obtains the lowest matched
MAE ($17.44^\circ$) and the highest $F_1@15^\circ$ (0.297), whereas
See-2-Sound obtains the highest $F_1@45^\circ$ (0.526). Thus, our advantage
is most evident at the stricter angular tolerance; recovering all
simultaneously active sources remains challenging.
We compare methods with defined spatial channel geometries. MMAudio's
mono output and Stable Audio 1.0's generic stereo do not provide a defined
absolute-azimuth reference and are not scored in this DoA comparison.
The reference inventory, activity mask, and channel-specific estimators
are specified in Appendix~\ref{sec:doa_app}.

\paragraph{Complementary spatial diagnostics.}
The original ILD and energy-ranking analyses remain complementary
source-level diagnostics (Appendix~\ref{sec:spatial_metric_app}); our ILD
correlation is 0.8958, compared with 0.3050 for the strongest baseline.
Neither these diagnostics nor mixture DoA establishes a proportional
relationship with perceptual source separation. The human study below
provides complementary evidence on complete mixtures.

\subsection{VLM-based and human preference studies}

The numerical metrics above test targeted aspects of the output, but spatial audio is ultimately perceptual. We therefore evaluate synchronized audio--video outputs using five criteria: overall audio--visual consistency, clarity of spatial changes, consistency with listener viewing direction and motion, spatial plausibility, and audio naturalness. Each criterion is rated on a 1--5 MOS scale, and evaluators also rank the candidate methods.

\paragraph{VLM-based evaluation.}
We use a two-stage blind protocol. First, rendered frames are paired with trajectory metadata, including camera positions, viewing directions, clip duration, and path statistics. GPT-5-mini uses these inputs to produce a structured description of the scene layout, listener motion, and spatial cues. Second, GPT-audio-mini evaluates each anonymized candidate soundtrack conditioned on this shared context and metadata. Candidate names are hidden behind labels such as \texttt{candidate\_a} to avoid method leakage; the full prompts are provided in the Appendix~\ref{sec:vlm_eval_app}. The original long-horizon VLM results are retained in Appendix~\ref{sec:long_vlm_results_app}.

\paragraph{Duration-matched comparison.}
We additionally compare all five methods using uniformly truncated 5\,s
clips, allowing See-2-Sound and OmniAudio to be evaluated under the same
duration constraint. In this automated evaluation,
Table~\ref{tab:vlm_eval_5s} shows that our method obtains the highest mean
MOS (3.01), the lowest mean rank (2.02), and the highest Rank-1 rate
(45.5\%). The MOS differences are small, while the ranking measures more
clearly favor our method. This is a separate short-horizon VLM-based
comparison, not an additional human study; its protocol is discussed in
Appendix~\ref{sec:short_vlm_app}.

\begin{figure*}[t]
    \centering

    \begin{minipage}{1.0\textwidth}

        \centering
        \captionof{table}{Human evaluation of methods using MOS and ranking-based evaluation metrics.}
        \label{tab:human_eval_summary}
        \setlength{\tabcolsep}{4pt}
        \renewcommand{\arraystretch}{1.1}
        \begin{tabular}{lccccc}
            \toprule
            \textbf{Method} &
            \textbf{Mean MOS} $\uparrow$ &
            \textbf{Mean Rank} $\downarrow$ &
            \textbf{Rank-1} $\uparrow$ &
            \textbf{Top-2} $\uparrow$ &
            \textbf{Borda} $\uparrow$ \\
            \midrule
            Stable Audio 1.0 & 2.268 & 2.429 & 8.0\%  & 49.1\% & 1.571 \\
            MMAudio          & \underline{2.498} & \underline{2.286} & \underline{12.5\%} & \underline{58.9\%} & \underline{1.714} \\
            Ours             & \textbf{4.145} & \textbf{1.286} & \textbf{79.5\%} & \textbf{92.0\%} & \textbf{2.714} \\
            \bottomrule
        \end{tabular}
    \end{minipage}

    \vspace{0.8em}

    \begin{minipage}[t]{0.49\textwidth}
        \centering
        \includegraphics[width=\linewidth]{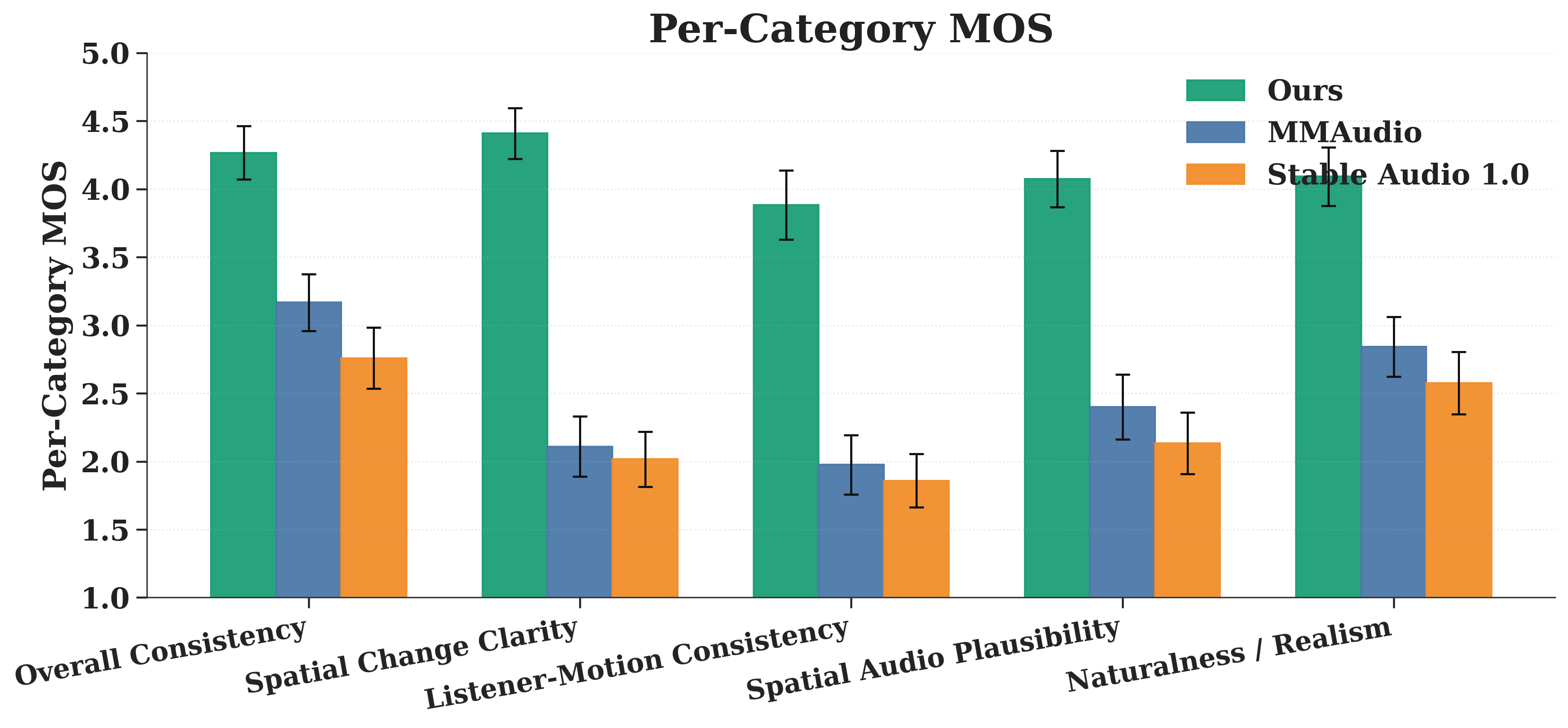}
        \captionof{figure}{Per-category MOS comparison across methods.}
        \label{fig:per_cat_mos_human}
    \end{minipage}
    \hfill
    \begin{minipage}[t]{0.49\textwidth}
        \centering
        \includegraphics[width=\linewidth]{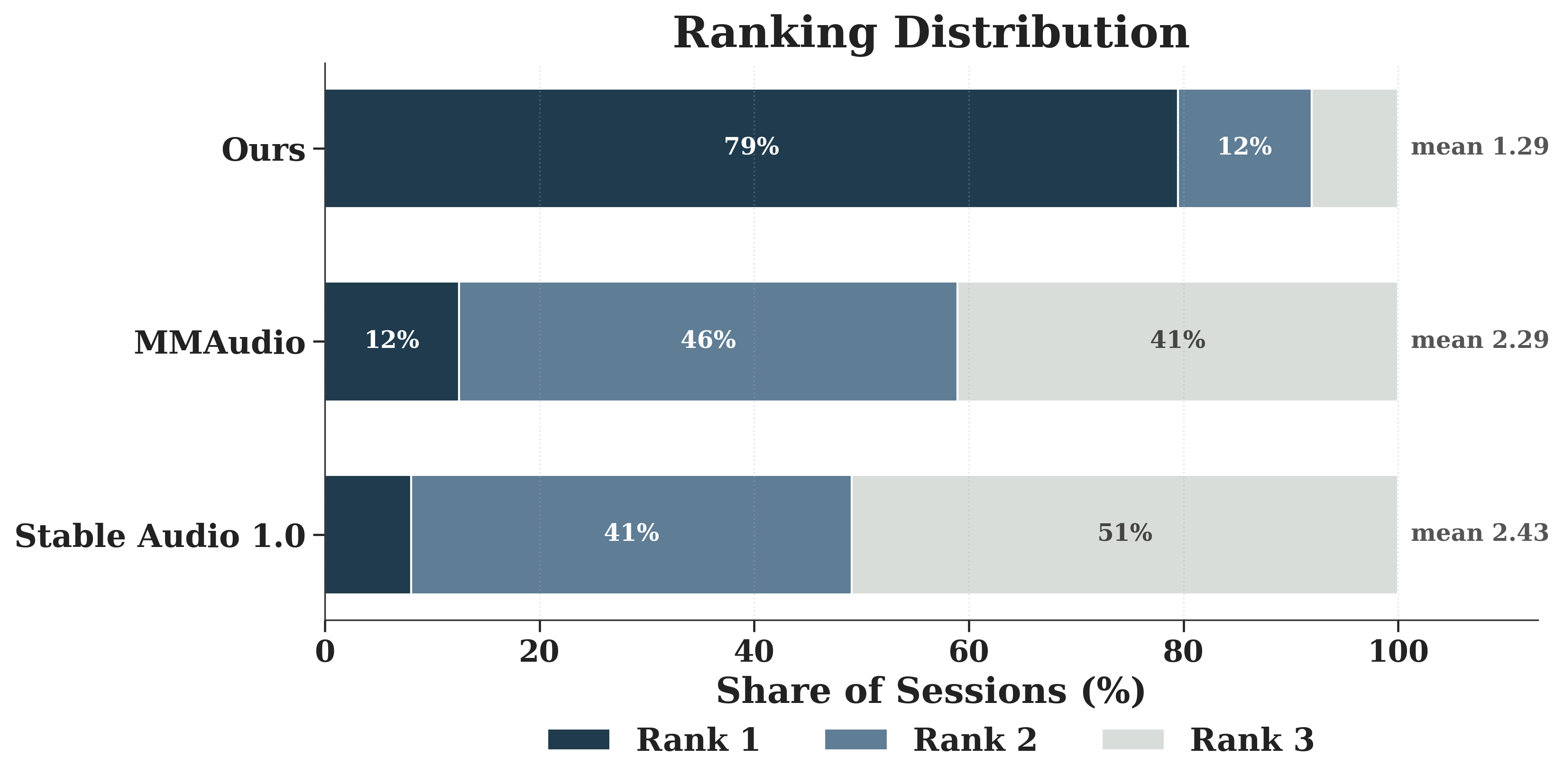}
        \captionof{figure}{Rank distribution comparison across methods.}
        \label{fig:rank_human}
    \end{minipage}

\end{figure*}

\paragraph{Human evaluation.}
We also conduct a blind human study on 20 generated scenes with 20 participants. For each scene, participants view the same rendered 3D trajectory video and evaluate three anonymized audio tracks: our method, Stable Audio 1.0, and MMAudio. They provide MOS ratings under the same five criteria and rank the three candidates from best to worst. Table~\ref{tab:human_eval_summary}, Figure~\ref{fig:rank_human}, and Figure~\ref{fig:per_cat_mos_human} show that our method is strongly preferred, with mean MOS 4.145, Rank-1 preference 79.5\%, and Top-2 preference 92.0\%. The strongest baseline reaches mean MOS 2.498, Rank-1 preference 12.5\%, and Top-2 preference 58.9\%, giving gains of 1.647 MOS points, 67.0 percentage points in Rank-1, and 33.1 percentage points in Top-2. This gap suggests that listeners value source grounding and motion-dependent acoustic rendering even when baselines generate plausible sound content.

\subsection{Qualitative spatial analysis}

Figure~\ref{fig:ild_qual_1} illustrates why the quantitative spatial scores improve. The rendered ILD follows the trend predicted by the listener--source geometry, and the channel-wise spectrograms show clear left--right changes as the listener moves. Figures~\ref{fig:sound_20s} and~\ref{fig:sound_40s} provide synchronized examples over short and long trajectories. Baseline methods often produce plausible audio events, but their waveforms exhibit weaker motion-dependent variation because the sound is not anchored to persistent 3D sources. In contrast, our output changes as the listener approaches, passes, or turns relative to the sources.

\subsection{Runtime and scalability}
\label{sec:runtime_scalability}
The system is an offline pipeline, and its constructed audible world can
be cached for subsequent listener trajectories. Holding the reconstructed
world fixed, source setup and acoustic propagation with 32 sources take
21.36\,s for a 5\,s trajectory and 184.14\,s for a 160\,s trajectory. At a fixed 20\,s duration, increasing
the source cap from 1 to 128 changes runtime from 22.32\,s to 92.10\,s.
Appendix~\ref{sec:cached_scaling_app} reports the complete sweeps and
reported timing variability, separately from end-to-end construction
costs. A controlled moving-source test with supplied source and listener
trajectories also renders 32 moving sources over 20\,s in 106.09\,s
(Appendix~\ref{sec:moving_sources_app}); this tests the rendering stage,
not automatic dynamic-world construction.

\section{Ablation Studies}
\label{sec:ablation}

We ablate the main components of the pipeline to test which parts are responsible for semantic quality and spatial consistency. The original component-removal ablations use localized per-source renders for spatial diagnostics and final full-scene soundtracks for semantic metrics. The additional heuristic and module-replacement experiments are reported separately with their own full-model references.

\paragraph{Trajectory length.}
Table~\ref{tab:ablation_short_long} compares 20-second and 40-second trajectories. The spatial scores remain stable across both settings, indicating that the method preserves source geometry over extended motion rather than only matching local, short-window cues.

\begin{table*}[t]
\centering
\scriptsize
\setlength{\tabcolsep}{3.5pt}
\resizebox{\textwidth}{!}{
\begin{tabular}{lcccc|cccc}
\toprule
& \multicolumn{4}{c|}{\textbf{Spatial metrics}} 
& \multicolumn{4}{c}{\textbf{Semantic metrics}} \\
\cmidrule(lr){2-5}\cmidrule(l){6-9}
\textbf{Variant}
& \textbf{ILD} $\uparrow$
& \textbf{ILD} $R^2$ $\uparrow$
& \textbf{Pair Acc.} $\uparrow$
& \textbf{Spearman} $\uparrow$
& \textbf{CLAP} $\uparrow$
& \textbf{IB-Text} $\uparrow$
& \textbf{IB-Image} $\uparrow$
& \textbf{Caption} $\uparrow$ \\
\midrule
Full model
& \textbf{0.888} & \textbf{0.807} & \textbf{0.823} & \textbf{0.758}
& 0.298 & \textbf{0.191} & \underline{0.179} & \textbf{0.301} \\
No acoustic simulation
& 0.007 & $\ll 0$ & 0.480 & 0.035
& \textbf{0.328} & 0.176 & 0.177 & \underline{0.275} \\
No agentic acoustic parameters
& \underline{0.881} & \underline{0.753} & 0.690 & 0.581
& 0.287 & 0.172 & 0.176 & 0.204 \\
No semantic decomposition
& -- & -- & -- & --
& \underline{0.324} & \underline{0.177} & 0.126 & 0.246 \\
No geometry-aware placement
& 0.033 & $\ll 0$ & 0.565 & 0.154
& 0.290 & 0.183 & 0.177 & 0.267 \\
No foreground/background separation
& 0.872 & 0.342 & \underline{0.729} & \underline{0.706}
& 0.287 & 0.164 & \textbf{0.185} & 0.221 \\
No layered 3D reconstruction
& -0.043 & $\ll 0$ & 0.561 & 0.160
& 0.187 & 0.119 & 0.114 & 0.187 \\
Scene-level TangoFlux~\cite{hung2024tangoflux}
& -- & -- & -- & --
& 0.225 & 0.158 & 0.172 & 0.190 \\
\bottomrule
\end{tabular}}
\caption{
World-sound ablation study. Spatial metrics are computed from localized per-source renders using the original scene geometry and a fixed ILD calibration. Semantic metrics are computed from the final full-scene soundtrack. ``--'' denotes cases where localized spatial metrics are not applicable.
}
\label{tab:world_sound_ablation}
\end{table*}

\paragraph{Component ablations.}
Table~\ref{tab:world_sound_ablation} shows that spatial consistency is most sensitive to acoustic propagation and geometric grounding. Removing the acoustic simulation stage nearly eliminates ILD correlation and produces unstable proxy fits, confirming that text-conditioned sound alone does not create reliable listener-dependent directional cues. Removing geometry-aware placement or layered 3D reconstruction similarly degrades both ILD and energy-ranking metrics, showing that sources must be tied to the generated scene rather than placed with a listener-centric heuristic.

The ablation of agentic acoustic parameters retains strong ILD correlation but reduces energy-ranking performance, suggesting that scene-adaptive parameters mainly improve relative source strength and attenuation. Removing foreground/background separation weakens both semantic and spatial behavior, likely because a single-pass parse produces less reliable source structure. Finally, the semantic metrics show that global audio--text similarity and spatial consistency are not equivalent: several variants remain competitive under CLAP while failing on spatial metrics. This supports our core claim that audible world generation requires explicit scene structure, not only a strong audio generator. We refer readers to Appendix~\ref{sec:ablation_app} for details of ablation study settings.

\paragraph{Comparison with heuristic construction.}
We compare against two heuristic baselines that retain the reconstructed
scene mesh and the GSound renderer. The weak baseline assigns generic
sounds from four broad scene categories; the strong baseline matches
prompt keywords to a fixed sound-prompt bank using sentence-embedding
similarity and synthesizes the retrieved prompts with Stable Audio.
Both place sources randomly on the scene mesh and randomly sample acoustic
parameters within the same permitted ranges. Our full method scores
higher on all four semantic metrics (Table~\ref{tab:heuristic_eval}).
These comparisons test the complete construction strategy rather than
isolating any single agent; implementation details are provided in
Appendix~\ref{sec:heuristic_app}.

\begin{table}[t]
\centering
\small
\setlength{\tabcolsep}{7pt}
\caption{Semantic comparison with heuristic world-sound construction.
Higher is better. This controlled comparison is reported separately from
the original 80-scene benchmark and component-removal ablations.}
\label{tab:heuristic_eval}
\begin{tabular}{@{}lrrrr@{}}
\toprule
Method & CLAP $\uparrow$ & IB-Text $\uparrow$
       & IB-Image $\uparrow$ & Caption $\uparrow$ \\
\midrule
Weak heuristic & 0.121 & 0.129 & 0.087 & 0.129 \\
Strong heuristic & 0.225 & 0.154 & 0.169 & 0.270 \\
Full method & \textbf{0.303} & \textbf{0.196}
            & \textbf{0.188} & \textbf{0.301} \\
\bottomrule
\end{tabular}
\end{table}

\paragraph{Single- and multi-module replacement.}
We test the modular interface through controlled substitutions of the
audio, segmentation, VLM, and depth backends
(Appendix~\ref{sec:module_swaps_app}). Replacing the source-audio backend
with TangoFlux changes CLAP from 0.303 to 0.222 and ILD correlation from
0.889 to 0.796. Replacing MoGe-2 with Depth Anything V2 yields CLAP 0.316
and ILD correlation 0.854. Jointly replacing the VLM, segmentation, and
depth modules while retaining Stable Audio yields CLAP 0.323 and ILD
correlation 0.893. These results support interchangeability of the tested
components, with metric-dependent tradeoffs, rather than invariance to
backend choice. The depth substitution measures downstream behavior, not
depth accuracy, because ground-truth depth is unavailable.

\paragraph{Inventory and agent repeatability.}
Scenes contain, on average, 1.17 foreground labels, 4.20 background labels,
and one global ambience track. These semantic components are distinct
from the sampled emitters used by the renderer. Across five repeated
parses of each evaluated scene, foreground/background semantic-label
agreement is 74.2\%/73.3\%, and localized-versus-diffuse decision agreement
is 82.6\%/88.1\%. Only 0.869\% of predicted sources require parameter
clamping. Appendix~\ref{sec:agent_diagnostics_app} reports count
distributions, uncertainty, and two-run parameter variation. These are
repeatability and range-compliance diagnostics, not detection-accuracy
measurements.
\section{Conclusion and Limitations}

We presented Audible World Models, a training-free framework for adding spatially aware audio to generated 3D worlds. The central idea is to treat sound as part of the world representation: audible objects and ambient regions are parsed from the generated scene, grounded on reconstructed geometry, assigned dry audio, and rendered through physically motivated acoustic propagation. This design yields a listener-dependent audiovisual experience with persistent source identities. Our evaluation combines semantic alignment over 80 generated scenes with geometry-referenced spatial tests: isolated-source DoA MAE is $6.91^\circ$, and mixture DoA reaches $17.44^\circ$ matched MAE and $F_1@15^\circ=0.297$. Human evaluation further increases mean MOS from 2.498 to 4.145 and Rank-1 preference from 12.5\% to 79.5\%. These results identify explicit world-space source grounding as the paper's central contribution.

The automatic construction pipeline currently produces static geometry
and fixed source placements. The supplied-trajectory experiment in
Appendix~\ref{sec:moving_sources_app} demonstrates moving-source rendering,
but does not infer object motion, reconstruct dynamic geometry, or establish
action--sound synchronization. The system is offline rather than real-time;
training-free operation does not remove the cost of world construction.
Audio fidelity depends on the audio backend, and acoustic realism depends
on reconstructed geometry and estimated parameters. Our geometric DoA
references test consistency with the constructed world, not recovery of
real-world source positions or exact material properties. Dynamic world
construction and more faithful geometry and material estimation remain
future work, including integration with long-horizon world models such as
Lyra 2.0~\cite{shen2026lyra2}.

\begin{ack}
We sincerely thank the reviewers for their valuable feedback. We gratefully acknowledge support from the Dolby ATG Research Award. We also thank NVIDIA for providing computational resources through the NVIDIA Academic Grant. The authors declare no competing interests.
\end{ack}

\bibliographystyle{plain}
\bibliography{egbib}

\appendix

\addcontentsline{toc}{part}{Appendix}
\section*{Appendix Contents}
\etocsettocstyle{}{}
\localtableofcontents
\clearpage

\section{Implementation Details of Each Ablation}
\label{sec:ablation_app}

For clarity and reproducibility, we summarize the exact implementation of each ablated variant used in our experiments.

\paragraph{No acoustic simulation.}
This variant keeps the same source set and 3D source positions as the full model, but bypasses the GSound propagation renderer entirely. Each localized source is evaluated using its original generated stereo waveform, resampled to the target sample rate, and scaled only by the source-level gain determined from its volume and source-power parameters. As a result, this ablation removes reflections, diffraction, transmission, and geometry-dependent listener-motion effects from the rendering stage.

\paragraph{No agentic acoustic parameters.}
This variant keeps the full geometry-aware placement and acoustic propagation pipeline, but replaces all per-source acoustic parameters with a fixed global preset. Specifically, reflectivity and scattering are clamped to constant values, and all sources share the same volume, source radius, and source power. This ablation isolates the effect of scene- and source-dependent acoustic parameter estimation.

\paragraph{No semantic decomposition.}
This variant bypasses per-label source generation and instead synthesizes a single scene-level soundtrack from the global scene prompt using Stable Audio 1.0. This setting reduces the pipeline to a text-to-audio baseline without explicit semantic source decomposition, and therefore matches the Stable Audio 1.0 comparison setting used in our experiments.

\paragraph{No geometry-aware placement.}
This variant keeps the same source waveforms and acoustic parameters as the full model, but discards the reconstructed scene-aware source locations. Instead, each source is placed on a fixed ring of radius $3\,\mathrm{m}$ around the listener, with azimuths evenly spaced from $-150^\circ$ to $150^\circ$ while preserving the original source height. This ablation directly tests whether geometry-grounded source placement is necessary for directional correctness.

\paragraph{No foreground/background separation.}
This variant starts from a single-pass panorama parse rather than the layered foreground/background decomposition used in the full pipeline. Sources whose labels are absent from the single-pass parse are removed. The remaining sources are treated as background-style scene objects: foreground sources inherit mean background acoustic parameters and are resampled on the corresponding semantic meshes. This ablation isolates the contribution of explicit foreground/background separation to source discovery, placement, and acoustic parameter assignment.

\paragraph{No layered 3D reconstruction.}
This variant uses the same single-pass panorama parse as the previous ablation, but further removes the explicit 3D lifting stage. Instead of reconstructing source positions in the 3D scene, each source is placed directly from its panorama bounding box. The bounding-box center determines azimuth and elevation, the source radius is chosen using a simple scene-scale heuristic, and duplicate instances of the same label receive small deterministic offsets in azimuth, elevation, and radius. This ablation tests whether the explicit layered 3D representation provides benefits beyond semantic parsing alone.

\paragraph{Scene-level alternative audio backend.}
This variant follows the global-track setup of the semantic-decomposition ablation, but regenerates the same scene prompt using TangoFlux~\cite{hung2024tangoflux} instead of the default audio backend. This experiment evaluates the sensitivity of the pipeline to the choice of text-to-audio generator and illustrates that the audio synthesis module can be substituted with alternative modern backends with minimal changes.

\subsection{Trajectory-length diagnostic}
We show spatial metric for different trjectory length in Table~\ref{tab:ablation_short_long}
\begin{table*}[t]
  \centering
  \caption{
  Spatial consistency under different trajectory lengths. ILD-based metrics measure agreement between the measured ILD curve and the fitted proxy ILD curve. Energy-ranking metrics compare the predicted source-energy ordering against the distance-based ordering. Higher is better for all metrics.
  }
  \label{tab:ablation_short_long}
  \setlength{\tabcolsep}{6pt}
  \renewcommand{\arraystretch}{1.15}
  \begin{tabular}{lcccc}
    \toprule
    \textbf{Method} &
    \textbf{ILD Corr.} $\uparrow$ &
    \textbf{ILD $R^2$} $\uparrow$ &
    \textbf{Pairwise Rank Acc.} $\uparrow$ &
    \textbf{Spearman Score} $\uparrow$ \\
    \midrule
    Ours (20 sec) & 0.8994 & 0.8110 & 0.8346 & 0.8063 \\
    Ours (40 sec) & 0.8875 & 0.8066 & 0.8232 & 0.7577 \\
    \bottomrule
  \end{tabular}
\end{table*}
\subsection{Heuristic world-sound construction}
\label{sec:heuristic_app}

We evaluate two baselines that replace world-aware source construction
with heuristic sound selection, source placement, and acoustic settings.
Their semantic results are reported in Table~\ref{tab:heuristic_eval}.

\paragraph{Weak heuristic.}
Each scene-generation prompt is assigned to one of four broad scene
categories, and generic default sounds associated with that category are
used as its semantic sound inventory.

\paragraph{Strong heuristic.}
We extract key terms from the scene-generation prompt and match them to a
fixed bank of sound-generation prompts. An LLM constructs the bank to
cover commonly occurring sounds. Sentence-transformer embeddings provide
cosine similarities between extracted keywords and bank labels; the
best-matched prompts are passed to Stable Audio to generate the semantic
soundtracks. The fixed resource is a prompt bank, not a set of measured
spatial source positions.

\paragraph{Shared geometry and rendering.}
For both baselines, sources are placed at random locations on the mesh
reconstructed by the full pipeline. Acoustic parameters are randomly
sampled within the same clamping ranges used by the full method, and
GSound remains the renderer. Spatial scores are not reported because the
heuristics do not infer semantically corresponding world-space source
positions; omitted scores should not be interpreted as zero error.
Because source selection, placement, and acoustic parameterization change
together, the comparison measures the complete construction strategy.

\subsection{Controlled module replacement}
\label{sec:module_swaps_app}

We replace one module at a time while retaining unrelated intermediate
states and downstream settings. The four substitutions replace the
Stable Audio source generator with TangoFlux~\cite{hung2024tangoflux},
SAM2 \cite{ravi2024sam} with ZIM \cite{kim2025zim} ViT-L, the GPT-based VLM with Claude Sonnet 4, and the
MoGe-2 \cite{wang2026moge} depth backend with Depth Anything V2 \cite{yang2024depth}. We additionally replace the
VLM and segmentation modules together, and then also replace depth.
The joint variants use Claude Opus 5 and retain Stable Audio, separating
these perception/geometry changes from a simultaneous audio-backend change.

\begin{table}[t]
\centering
\small
\setlength{\tabcolsep}{4pt}
\caption{Controlled single- and multi-module replacement. All metrics
are higher-is-better. Caption is the Audio Flamingo 2 caption-similarity
metric (AF2 in the experiment record). Joint variants retain Stable Audio.
These results have their own reported full-model reference and do not
replace the original benchmark or component-removal results.}
\label{tab:module_swaps}
\begin{tabular}{@{}lrrrrrr@{}}
\toprule
Variant & CLAP & IB-T & IB-I & Caption & ILD Corr. & ILD $R^2$ \\
\midrule
Full model & 0.303 & 0.196 & 0.188 & 0.301 & 0.889 & 0.812 \\
\midrule
Audio $\to$ TangoFlux & 0.222 & 0.167 & 0.189 & 0.205 & 0.796 & 0.740 \\
Segmentation $\to$ ZIM & 0.280 & 0.180 & 0.204 & 0.255 & 0.885 & 0.791 \\
VLM $\to$ Sonnet 4 & 0.300 & 0.171 & 0.222 & 0.224 & 0.875 & 0.776 \\
Depth $\to$ DA-V2 & 0.316 & 0.185 & 0.215 & 0.336 & 0.854 & 0.746 \\
\midrule
VLM + segmentation & 0.321 & 0.180 & 0.209 & 0.279 & 0.876 & 0.788 \\
VLM + segmentation + depth & 0.323 & 0.184 & 0.205 & 0.274 & 0.893 & 0.806 \\
\bottomrule
\end{tabular}
\end{table}

Table~\ref{tab:module_swaps} shows that the tested substitutions remain
compatible with the pipeline but affect metrics differently. For example,
the audio substitution reduces CLAP and Caption, while IB-Image changes
from 0.188 to 0.189. The depth substitution increases CLAP and Caption
but decreases ILD correlation and $R^2$. Thus, these results demonstrate
modular execution with quality tradeoffs, not statistical equivalence
between backends. Without ground-truth depth, the depth experiment cannot
establish which reconstruction is geometrically more accurate.
The source-level audio substitution here is distinct from the original
\emph{scene-level} TangoFlux ablation, which synthesizes a single global
track without semantic source decomposition.

\section{Video Rendering and Baseline Audio Generation}
\label{sec:baselines}
\subsection{3D World Rendering}
For each scene session, we first render a canonical silent video clip. This clip serves as the visual input to video-conditioned baselines and as the carrier stream for all final audio--video composites. Each session directory contains a reconstructed scene representation and is paired with a predefined camera-trajectory text file. We parse the trajectory, uniformly sample frame timestamps over the target clip duration, and linearly interpolate camera positions along the path. In our rendering script, the short benchmarks use a $20$\,s clip and the long benchmarks use a $40$\,s clip. Frames are rendered at $15$\,fps with resolution $960\times540$ and a $75^\circ$ field of view. The renderer then outputs one RGB frame for each sampled camera pose.

\subsection{Baselines Generation}

\paragraph{MMAudio.}
MMAudio is conditioned directly on the rendered silent video. For each session, we provide the silent clip as input, together with the selected model variant, the target duration, the classifier-free guidance strength, the number of diffusion steps, and a fixed random seed. In our experiments, we use the \texttt{large\_44k\_v2} variant. When available, we also parse the original scene-generation prompt and pass it to MMAudio as an optional text condition. MMAudio produces both a synthesized audio waveform and an audio--video composite obtained by attaching the generated waveform to the input clip, which we use for qualitative inspection.

\paragraph{Waveform-only baselines and audio--video compositing.}
For methods that generate audio but do not directly output a composite video, we reuse the same silent clip and remux the generated audio onto it. This category includes our acoustic renderings, Stable Audio 1.0, and other waveform-only baselines such as TangoFlux. We combine video and audio with a single \texttt{ffmpeg} call that maps the video stream from rendered world video and the audio stream from the generated waveform, copies the original video bitstream without re-encoding, encodes the audio as AAC, truncates the output to the desired clip length, and stops at the shorter stream if needed. If a generated waveform contains non-finite values, we first sanitize it by replacing NaNs/Infs before muxing. This ensures that all methods are compared using exactly the same visual content.

\paragraph{Stable Audio 1.0}
We first rewrite the scene-generation prompt into an audio-oriented prompt using an LLM (gpt-5-mini), and then generate a single audio clip from the refined prompt. We use the released \texttt{StableAudioPipeline}(\texttt{stabilityai/stable-audio-open-1.0}) for generating all the results for comparison with default number of inference steps and guidance scale. For video with sound tracks, we mux the generated sound track back onto the rendered world view video as mentioned above.

\paragraph{SEE-2-SOUND.}
SEE-2-SOUND is conditioned on the scene panorama rather than the rendered camera clip. For each session, we provide the generated \texttt{panorama.png} to the released SEE-2-SOUND inference wrapper and use the resulting sound track as the baseline output.

\paragraph{OmniAudio.}
OmniAudio is video-conditioned but expects a shorter input clip than our full rendered sequence. We therefore create a temporary $10$\,s version of \texttt{render\_silent.mp4} and pass it to OmniAudio's released inference pipeline using the FOA model configuration.
\section{Agent Prompts and Structured Outputs}
\label{sec:vlm_prompts_node}

All agent calls use the generated panorama and the original scene prompt as context. The original prompt is treated as a weak prior: when the prompt and generated image disagree, the visible/generated scene is prioritized. We request structured JSON-like outputs so that downstream modules can consume labels, masks, sound descriptions, and acoustic parameters without free-form parsing.

\subsection{Foreground sound-source parser}
\label{sec:foreground_prompt_app}
The foreground parser is applied in two passes. The first pass identifies salient unobscured objects. The second pass operates after conceptual removal of the first layer and identifies remaining localized foreground entities. We restrict the second pass to compact object-like entities to avoid assigning global regions such as sky, fog, atmosphere, or general lighting to foreground layers.

\subsubsection*{Task Prompt}
\begin{PromptBlock}[prompt/task]
Given the panoramic image above, please complete the following tasks:
1) Identify up to ONE unobscured and most salient foreground object categories for FOREGROUND LAYER 1 (fg1).
   - Return 1 labels only. Do not return more than one labels.
2) Now imagine fg1 objects are removed/inpainted. Identify up to One remaining salient FOREGROUND object categories for FOREGROUND LAYER 2 (fg2).
   - Return 1 labels only. Do not return more than one labels.
   - fg2 must be discrete foreground entities (countable/localized), not global scene regions.
   - Do NOT include sky/atmosphere/non-object descriptors (sky, clouds, sun, moon, stars, haze, fog, horizon, background).
   - trees/buildings/water/roads ARE allowed only if clearly localized foreground entities (e.g., 'nearby tree trunk', 'foreground building facade', 'boat').
3) Provide an inpainting prompt for removing fg1 objects, and another for removing fg2 objects.
   IMPORTANT (CRITICAL FOR QUALITY):
   - Each inpainting prompt must ONLY describe what should appear in the removed/masked region.
   - The prompt MUST NOT mention or name the removed objects (do not include their nouns at all).
   - Describe plausible continuation of the surrounding content (materials/structures/lighting/perspective).
   - Keep it SHORT: ONE sentence, <= 20 words, <= 140 characters.
   - Do NOT use long comma-separated lists.
4) Determine whether the scene is indoor or outdoor.
5) For any fg labels likely to produce sound, provide per-label:
   - sound description, give descriptions that descriptive and informational but always keep in mind of the context prompt., and
   - whether the sound should be DIRECTIONAL (true/false).
   - if there's animal/instrument/engine/etc that can be reasonably assumed to be making sound, label it and give a sound description even if it's not salient visually.
   Directional=true for localized point-like sources (e.g., waterfall, car engine, bird near camera).
   Directional=false for diffuse/ambient sources (e.g., wind, crowd murmur, broad rustling).
\end{PromptBlock}

\subsubsection*{Schema Hint}
\begin{PromptBlock}[prompt/schema]
Return ONLY a JSON object with this schema:
{
  "classes": "indoor" | "outdoor",
  "fg1": {
    "labels": ["label1"],
    "inpainting_prompt": "ONE sentence, <= 20 words, <= 140 chars; describes ONLY what should appear in the masked region; MUST NOT mention removed objects"
  },
  "fg2": {
    "labels": ["label1"],
    "inpainting_prompt": "ONE sentence, <= 20 words, <= 140 chars; describes ONLY what should appear in the masked region; MUST NOT mention removed objects"
  },
  "sound_descriptions": {
    "label": {
      "description": "sound description, give descriptions that descriptive and informational but always keep in mind of the context prompt.",
      "directional": true
    }
  }
}
Rules:
- labels must be concrete nouns (or short noun phrases) suitable for segmentation.
- Avoid duplicates and synonyms; pick canonical labels.
- fg1.labels must contain 1 to 2 labels.
- fg2.labels must contain 0 to 2 labels.
- fg2.labels MUST NOT contain sky/atmosphere/non-object descriptors.
- If fg2 has no objects, set fg2.labels=[] but still provide fg2.inpainting_prompt.
\end{PromptBlock}

\subsubsection*{Full Prompt Wrapper Sent to the LLM}
\begin{PromptBlock}[prompt/wrapper]
{OPENAI_TASK_PROMPT}

Panorama generation prompt, you should perform task by keeping in mind this prompt, especially for generating sound description:
'''{pano_prompt}'''

{OUTPUT_SCHEMA_HINT}
\end{PromptBlock}

\subsection{Background and ambience parser}

After foreground removal, a second parser identifies broader sound-producing regions and global ambience. This pass is designed to recover off-screen or distributed sources that a short rendered video might miss, such as trees, rivers, markets, ocean surf, city traffic, or room tone.
\subsubsection*{Task Prompt}
\begin{PromptBlock}[prompt/task]
Given the panoramic image above, please complete the following tasks:
1. Identify the semantic labels in the image that will generate sound. For each label:
   - sound description, give descriptions that descriptive and informational but always keep in mind of the context prompt.
   - specify whether the sound is DIRECTIONAL (true/false)
     (Directional=true for localized point-like sources; false for diffuse/ambient sounds).
2. For each label, provide likely bounding box locations in a json file.
3. If there should be a background/ambience sound for the image, provide:
   - sound description, give descriptions that descriptive and informational but always keep in mind of the context prompt.
   - if there's animal/instrument/engine/etc that can be reasonably assumed to be making sound, label it and give a sound description even if it's not salient visually.
   - For animals, the description should just be their normal sound (e.g., 'birdsong', 'cow mooing')
   - directional (usually false for ambience)
\end{PromptBlock}

\subsubsection*{Alignment Prompt}
\begin{PromptBlock}[prompt/align]
Important context: the panorama was generated from the following prompt.
- Use the prompt as semantic prior for what sounds/objects are likely and how to describe them.
- If the prompt conflicts with the image, ALWAYS follow the image.
- Do NOT invent objects not supported by the image.

Label requirements:
- Labels must be segmentable concrete nouns (e.g., 'trees', 'waves', 'traffic', 'waterfall', 'birds'),
  NOT abstract phrases.
- Avoid synonyms for the same thing (pick one canonical label).

Bounding boxes:
- Prefer PIXEL coordinates [x1,y1,x2,y2] for the provided image size.
- If you output normalized coordinates, set coord_type='normalized'.
- Scores should be in [0,1].
\end{PromptBlock}

\subsubsection*{Schema Hint}
\begin{PromptBlock}[prompt/schema]
Return ONLY a JSON object with this schema:
{
  "coord_type": "pixel" | "normalized",
  "sound_objects": [
    {
      "label": "string",
      "sound_description": "string",
      "directional": true/false,
      "boxes": [
        {"x1": number, "y1": number, "x2": number, "y2": number, "score": number}
      ]
    }
  ],
  "background_sound": {"present": true/false, "description": "string"}
}
Notes: If coord_type='pixel', x in [0,{img_w}] and y in [0,{img_h}]. If coord_type='normalized', coords are in [0,1] and will be scaled.
\end{PromptBlock}

\subsubsection*{Full Prompt Wrapper Sent to the LLM}
\begin{PromptBlock}[prompt/wrapper]
{OPENAI_TASK_PROMPT}
{OPENAI_ALIGNMENT_PROMPT}
Image size: width={img_w}, height={img_h}
Panorama generation prompt:
'''{pano_prompt}'''
you should perform task by keeping in mind this prompt, especially for generating sound description

{schema_hint}
\end{PromptBlock}

\subsection{Acoustic parameter agent}
\label{sec:acoustic_params_app}

The acoustic parameter agent proposes conservative scene-level and source-level settings. These values are clamped before simulation and are intended to capture qualitative acoustic character, not exact material recovery.
\begin{PromptBlock}[prompt/task]
You are tuning acoustic simulation parameters for a ray-based propagation renderer.
Return ONLY JSON using the given schema.

Scene type: {classes}
Panorama prompt (prior): {pano_prompt}
Scene generation prompt (hint): {scene_prompt_hint}

Sound labels and descriptions:
{json.dumps(per_label, indent=2, ensure_ascii=False)}

Directional flags (true=localized point-like, false=diffuse):
{json.dumps(per_label_dir, indent=2, ensure_ascii=False)}

Approx bbox counts (proxy for how many sources are placed per label):
{json.dumps(counts, indent=2, ensure_ascii=False)}

Per-label source location summary (distance to origin [0,0,0]):
{json.dumps(source_loc_summary, indent=2, ensure_ascii=False)}

Preferred ranges (must follow):
- volume: 40..90
- radius_ratio: 0.0001..0.05
- power_db: 65..80

Default source presets (baseline; modify based on scene and distances):
{json.dumps(default_source_presets, indent=2, ensure_ascii=False)}

Schema:
{
  "global": {
    "reflectivity": number (0..1),
    "scattering": number (0..1),
    "render_volume": number (0.1..2),
    "ir": number (0.5..6),
    "toggles": {
      "direct": bool, "specular": bool, "diffuse": bool,
      "diffraction": bool, "transmission": bool, "air": bool
    }
  },
  "source_presets": {
    "background_ambience": {"volume": number(40..90), "radius_ratio": number(0.0001..0.05), "power_db": number(65..80)},
    "bg_label_dir":        {...},
    "bg_label_diff":       {...},
    "fg_dir":              {...},
    "fg_diff":             {...}
  },
  "per_label_overrides": {
    "label": {"volume": number(40..90), "radius_ratio": number(0.0001..0.05), "power_db": number(65..80)}
  }
}

Guidelines:
- Outdoor scenes: lower reflectivity, higher scattering than indoor.
- Directional sources: smaller radius_ratio, higher power_db than diffuse.
- Background ambience: low power_db and larger radius_ratio.
- [IMPORTANT!] Use distance-to-origin as a volume and power_db setting hint:
  labels farther from origin should be boosted (higher volume, power_db, and slightly larger radius_ratio)
  via source_presets and/or per_label_overrides so they remain audible.
- Use real-world scene priors from the scene generation prompt:
  set relative volume/power_db/radius_ratio BETWEEN labels to match how the scene usually sounds.
  The ratios between labels matter more than absolute values.
- Keep values conservative to avoid clipping.
- Try to keep every label audible in the mix.
\end{PromptBlock}

\subsection{Source inventory and agent diagnostics}
\label{sec:agent_diagnostics_app}

\paragraph{Semantic inventory.}
Table~\ref{tab:source_inventory} summarizes foreground labels, background
labels, and global ambience tracks. The two-layer foreground decomposition
limits the number of separately segmented, removed, inpainted, and
reconstructed foreground layers; it does not limit the complete audible
inventory to two sounds. On average, a scene contains 5.37 scene-grounded
labels (1.17 foreground and 4.20 background), plus one global ambience
track. Background labels have their own descriptions, dry assets, visual
support, and spatial grounding, whereas global ambience is a separate
non-localized component. A semantic label may be represented by multiple
sampled acoustic emitters, so label counts differ from rendering-source
counts.

\begin{table}[t]
\centering
\small
\setlength{\tabcolsep}{5pt}
\caption{Distribution of semantic audio components per scene. These
counts describe labels/tracks, not sampled acoustic emitters.}
\label{tab:source_inventory}
\begin{tabular}{@{}lrrrrrrr@{}}
\toprule
Component & Mean & Median & SD & Min & 25th pct. & 75th pct. & Max \\
\midrule
Foreground labels & 1.17 & 1 & 0.55 & 0 & 1 & 1.25 & 2 \\
Background labels & 4.20 & 4 & 1.53 & 1 & 3 & 5 & 8 \\
Ambience tracks & 1.00 & 1 & 0.00 & 1 & 1 & 1 & 1 \\
\bottomrule
\end{tabular}
\end{table}

\paragraph{Repeated semantic parsing.}
We run the GPT-5-mini parser five times for each evaluated scene with the
same prompt. Labels are embedded with the \texttt{all-MiniLM-L6-v2}
sentence-transformer model; labels whose cosine similarity exceeds 0.7
are treated as semantic matches. We also measure
agreement of the directional flag, which denotes localized-versus-diffuse
source behavior rather than a continuous source azimuth.
Table~\ref{tab:parser_repeatability} reports mean agreement, scene standard
deviation, and 95\% bootstrap confidence intervals. These results quantify
repeatability, not accuracy against human-annotated semantic labels.

\begin{table}[t]
\centering
\small
\setlength{\tabcolsep}{7pt}
\caption{Parser repeatability over five runs per evaluated scene. Entries
are mean $\pm$ scene SD; bracketed values are 95\% bootstrap confidence
intervals. All values are percentages.}
\label{tab:parser_repeatability}
\begin{tabular}{@{}lcc@{}}
\toprule
Parser & Semantic-label agreement & Directional-flag agreement \\
\midrule
Foreground & $74.2 \pm 16.5$ & $82.6 \pm 24.5$ \\
           & $[67.5,81.5]$ & $[71.4,92.5]$ \\
Background & $73.3 \pm 12.1$ & $88.1 \pm 10.0$ \\
           & $[68.1,78.3]$ & $[83.7,92.2]$ \\
\bottomrule
\end{tabular}
\end{table}

\paragraph{Acoustic parameter repeatability and clamping.}
The acoustic-agent prompt specifies permitted parameter ranges before
prediction. In the reported runs, 0.869\% of predicted sound sources
contain an out-of-range parameter and therefore require clamping.
Table~\ref{tab:parameter_repeatability} summarizes mean absolute parameter
changes between two runs of the same session, both in the parameter's
native scale and relative to its permitted span.

\begin{table}[t]
\centering
\small
\setlength{\tabcolsep}{7pt}
\caption{Acoustic parameter variation between two runs of the same
session, averaged over the reported sessions. The final column normalizes
the absolute change by the parameter's allowed span.}
\label{tab:parameter_repeatability}
\begin{tabular}{@{}lrr@{}}
\toprule
Parameter & Mean absolute change & Mean \% of allowed span \\
\midrule
Reflectivity & 0.0217 & 2.17\% \\
Scattering & 0.0688 & 6.88\% \\
Render volume & 0.0210 & 1.11\% \\
Source volume & 3.71 & 7.41\% \\
Source radius ratio & 0.00197 & 3.95\% \\
Source power & 1.13\,dB & 7.53\% \\
\bottomrule
\end{tabular}
\end{table}

\paragraph{Observed inventory emptiness.}
No reported scene has both an empty foreground inventory and an empty
background inventory. This observation and the clamping rate diagnose
empty outputs and range compliance; they do not measure missed-object
recall, correctness of predicted parameters, or an overall pipeline
success rate against annotated ground truth.

\section{Layer Decomposition and Mask Processing}
\label{sec:layer_mask_app}

\paragraph{Foreground layers.}
The foreground parser returns labels for segmentation and fill-only inpainting prompts for object removal. Fill-only prompts describe the desired replacement content inside the mask rather than restating the removed object. This choice reduces semantic regrowth, where the inpainting model reconstructs the removed category because the prompt still mentions it. For example, when removing a fountain from a plaza, the fill prompt should describe ``continuous plaza paving and background architecture'' rather than ``remove the fountain from the plaza.''

\paragraph{Mask generation and cleanup.}
For each label, we obtain a candidate mask using the segmentation backend. We discard masks whose area is too small to be reliable or too large for a compact foreground object. For overlapping foreground masks, we keep the higher-confidence or more specific label and subtract it from lower-priority masks. For background masks, overlap is allowed when a region can plausibly support multiple sounds, but duplicated near-identical masks are merged.

\paragraph{Background parsing after foreground removal.}
The background parser operates on a foreground-removed panorama. This avoids a common failure mode in which salient objects dominate the parse and suppress broader ambience. For example, a single visible car should not prevent the model from identifying city traffic, crowd murmur, or street ambience in the rest of the panorama.

\paragraph{Sky handling.}
Sky and atmosphere are not treated as localized foreground sources. They are only used for visual scene composition and, when appropriate, for global ambience descriptions such as wind, rain, thunder, or distant outdoor ambience.

\section{Source Placement and Energy Normalization}
\label{sec:source_placement_app}

\subsection{Surface sampling}

Each localized source label is associated with a semantic mesh or world-sheet region. Given a triangular mesh with faces \(f\) and areas \(A_f\), we sample a face with probability
\begin{equation}
    p(f) = \frac{A_f}{\sum_{f'} A_{f'}}.
\end{equation}
For a selected triangle with vertices \(\mathbf{v}_0,\mathbf{v}_1,\mathbf{v}_2\), we sample barycentric coordinates uniformly and obtain a surface point \(\mathbf{p}\). If \(\mathbf{n}\) is the local unit normal, the source is offset slightly from the surface as
\begin{equation}
    \mathbf{x} = \mathbf{p} - \mathbf{n}(\delta + \epsilon),
\end{equation}
where \(\delta\) is a mesh-scale-dependent offset and \(\epsilon\) is small random jitter. This offset prevents degenerate source--surface coincidence in the acoustic renderer. For diffuse labels, we sample multiple points over the semantic support region. For compact labels, we use fewer samples and optionally collapse nearby samples to avoid over-counting a single object.

\subsection{Source-count normalization}

When a semantic label is represented by \(N\) sampled source points, naive playback can make that label louder simply because more points were sampled. We therefore normalize source energy across all points belonging to the same label. If \(P_0\) is the nominal source power and \(v_0\) is the nominal volume, each sampled point receives
\begin{equation}
    P = P_0 - 10\log_{10}(N), \qquad
    v = \frac{v_0}{\sqrt{N}}.
\end{equation}
This keeps the aggregate emitted energy approximately stable as the number of samples changes. The normalization is especially important for distributed ambience, where a forest, crowd, or ocean region may be represented by many source points.

\subsection{Localized versus diffuse labels}

A localized label is intended to have a compact, inspectable world-space origin, such as a dog, vehicle, performer, fountain, or machine. A diffuse label is distributed across a larger visual support, such as wind in trees, crowd murmur, ocean surf, or market ambience. The locality attribute affects both the number of source samples and the acoustic radius assigned to each sample. Localized labels typically use smaller radii and fewer samples; diffuse labels use larger radii or multiple distributed samples.

\section{VLM Qualitative Evaluation Prompts}
\label{sec:vlm_eval_app}
\subsubsection*{Scene Summarization System Prompt}
\begin{PromptBlock}[prompt/system]
You summarize the shared visual context for a soundtrack evaluation study. The images come from one 3D scene and one fixed camera trajectory. Your job is to describe the visible scene content and the listener motion cues that a later audio judge should use. When explicit trajectory metadata is provided, treat it as the ground-truth listener motion and viewing direction. Use the images to connect that motion to visible objects and scene regions. Do not evaluate rendering quality. Do not invent sounds unless they are visually grounded.
\end{PromptBlock}

\subsubsection*{Scene Summarization User Prompt Template}
\begin{PromptBlock}[prompt/user]
Summarize the shared scene context from the attached images.

Explicit listener trajectory metadata:
{trajectory_text}

Return JSON only with this schema:
{
  "scene_summary": "brief visual description of the scene and main actions",
  "camera_trajectory_summary": "brief description of the camera path over time",
  "listener_motion_summary": "brief description of the implied listener orientation and motion",
  "salient_visible_events": ["short event or object cue", "short event or object cue"],
  "spatial_audio_cues": ["short cue relevant to spatial sound alignment", "short cue relevant to spatial sound alignment"],
  "uncertainties": ["brief uncertainty if any"]
}

Rules:
- Keep the summary concise and concrete.
- Focus on cues that matter for judging soundtrack alignment.
- If explicit trajectory metadata is provided, use it as the authoritative listener motion/view-direction signal.
- Use the images to identify which visible objects or scene regions align with that motion.
- If there is no clear uncertainty, return an empty list for "uncertainties".
\end{PromptBlock}

\subsubsection*{Independent Audio Evaluation System Prompt}
\begin{PromptBlock}[prompt/system]
You are running a human-study-style soundtrack evaluation. The video frames have already been summarized for you in text. Judge only the soundtrack against that shared scene context and apparent listener motion. Do not judge rendering quality or video quality. Be conservative and internally consistent.
\end{PromptBlock}

\subsubsection*{Independent Audio Evaluation User Prompt Template}
\begin{PromptBlock}[prompt/user]
Evaluate one candidate soundtrack independently.

Candidate soundtrack id: "{soundtrack_alias}"

Shared scene context:
{json_prompt_blob(scene_context)}

Explicit listener trajectory metadata:
{trajectory_text}

Rate the soundtrack on a 1-5 integer scale for:
1. overall_consistency: Overall consistency between the video and the sound
2. spatial_change_clarity: Clarity of spatial changes in the sound
3. listener_motion_consistency: How well the sound reflects the listener's viewing direction and motion
4. spatial_audio_plausibility: Plausibility of the spatial audio
5. naturalness_realism: Naturalness / realism of the audio

Important instructions:
- Judge this candidate independently. Do not compare it against any other candidate.
- Focus on the soundtrack, not the rendering quality.
- Use the explicit trajectory metadata as the ground-truth listener motion when it is provided.
- If no explicit trajectory metadata is provided, fall back to motion inferred from the frames.
- When rating spatial_change_clarity, consider whether spatial variation is noticeable and understandable.
- When rating listener_motion_consistency, consider whether audio changes match the listener orientation and movement implied by the trajectory metadata and camera trajectory summary.
- Keep justifications brief but specific.
- If the spatial effect is weak or unclear, score it low rather than assuming spatial motion.

Return JSON only with this schema:
{
  "soundtrack_id": "{soundtrack_alias}",
  "scores": {
    "overall_consistency": {"score": 1, "justification": "brief"},
    "spatial_change_clarity": {"score": 1, "justification": "brief"},
    "listener_motion_consistency": {"score": 1, "justification": "brief"},
    "spatial_audio_plausibility": {"score": 1, "justification": "brief"},
    "naturalness_realism": {"score": 1, "justification": "brief"}
  },
  "overall_comment": "brief overall comment"
}
\end{PromptBlock}

\subsubsection*{Final Ranking System Prompt}
\begin{PromptBlock}[prompt/system]
You are producing the final human-study-style ranking for soundtrack candidates. Each candidate was already judged independently. Use the shared scene context and the independent evaluations to rank the candidates best to worst overall. Focus on sound quality, clarity of spatial motion, and consistency with the scene and listener motion. Do not judge rendering quality.
\end{PromptBlock}

\subsubsection*{Final Ranking User Prompt Template}
\begin{PromptBlock}[prompt/user]
Rank {len(candidate_ids)} soundtrack candidates for the same scene from best to worst overall.

Shared scene context:
{json_prompt_blob(scene_context)}

Explicit listener trajectory metadata:
{trajectory_text}

Independent candidate evaluations:
{json_prompt_blob(independent_evaluations)}

Important instructions:
- Judge the final ranking using the shared scene context and the independent evaluations above.
- The scene content and camera trajectory are the same for all candidates.
- Treat the explicit trajectory metadata as the ground-truth listener motion when it is provided.
- Focus on sound quality, clarity of spatial motion, and consistency with the scene/listener motion.
- Produce a strict total ordering with no ties.
- Keep justifications brief but specific.

Return JSON only with this schema:
{
  "ranking": [
    {"rank": 1, "soundtrack_id": "candidate_x", "justification": "brief"},
    ...
  ],
  "overall_comment": "brief summary of why the top-ranked candidate won"
}
\end{PromptBlock}

\subsection{Original long-horizon VLM results}
\label{sec:long_vlm_results_app}
The following table and plots retain the original three-method long-horizon VLM study.
\begin{figure*}[t]
    \centering

    \begin{minipage}{1.0\textwidth}
        \centering
        \captionof{table}{Original long-horizon VLM-based comparison over 20--40\,s trajectories. Higher is better except mean rank. These are three-candidate results, distinct from the five-second comparison.}
        \label{tab:vlm_eval_summary}
        \setlength{\tabcolsep}{4pt}
        \renewcommand{\arraystretch}{1.1}
        \begin{tabular}{lccccc}
            \toprule
            \textbf{Method} &
            \textbf{Mean MOS} $\uparrow$ &
            \textbf{Mean Rank} $\downarrow$ &
            \textbf{Rank-1} $\uparrow$ &
            \textbf{Top-2} $\uparrow$ &
            \textbf{Borda} $\uparrow$ \\
            \midrule
            Stable Audio 1.0 & 2.93 & 2.33 & 16.7\% & 50.0\% & 1.67 \\
            MMAudio & \underline{2.99} & \underline{1.99} & \underline{29.5\%} & \underline{71.8\%} & \underline{2.01} \\
            Ours & \textbf{3.02} & \textbf{1.68} & \textbf{53.8\%} & \textbf{78.2\%} & \textbf{2.32} \\
            \bottomrule
        \end{tabular}
    \end{minipage}

    \vspace{0.8em}

    \begin{minipage}[t]{0.49\textwidth}
        \centering
        \includegraphics[width=\linewidth]{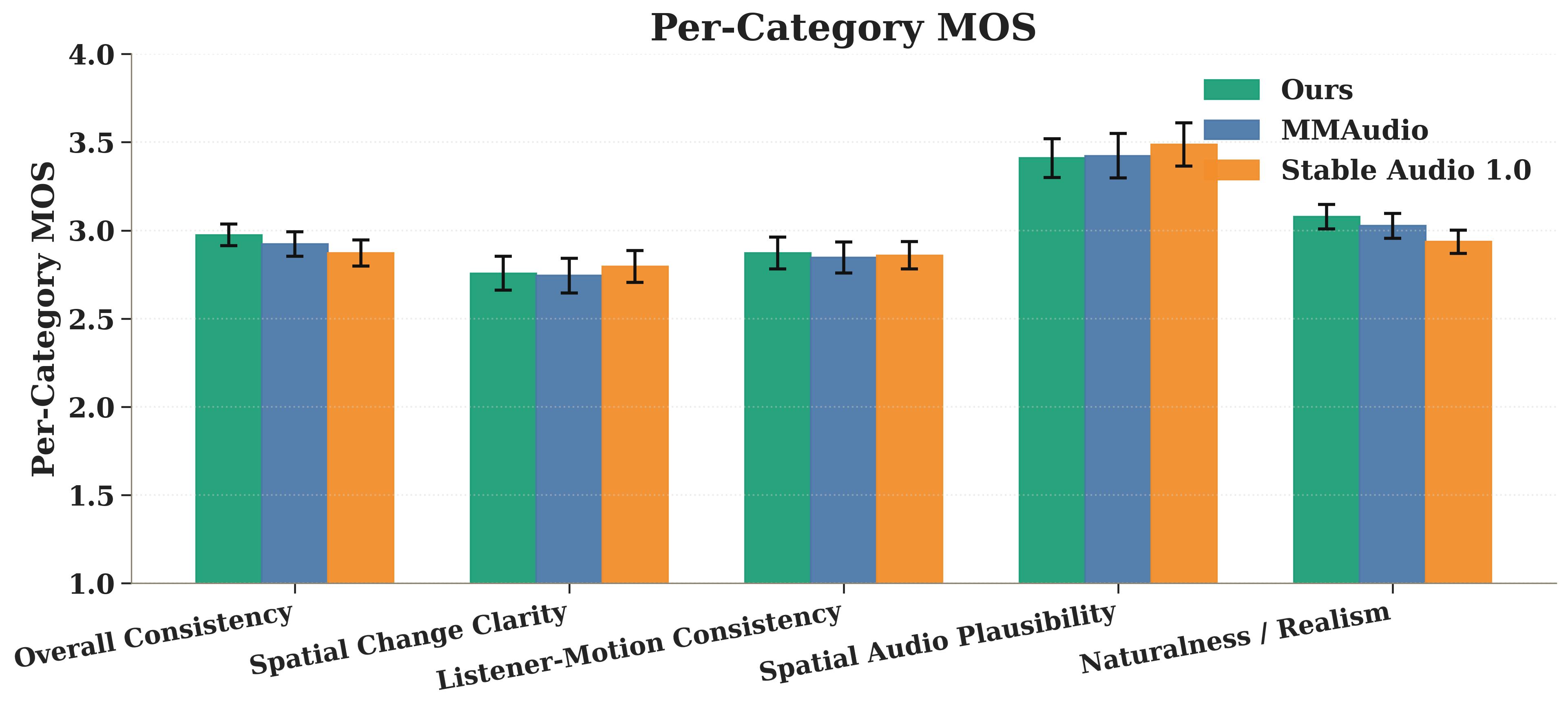}
        \captionof{figure}{Original long-horizon VLM per-category MOS for three methods.}
        \label{fig:per_cat_mos_vlm}
    \end{minipage}
    \hfill
    \begin{minipage}[t]{0.49\textwidth}
        \centering
        \includegraphics[width=\linewidth]{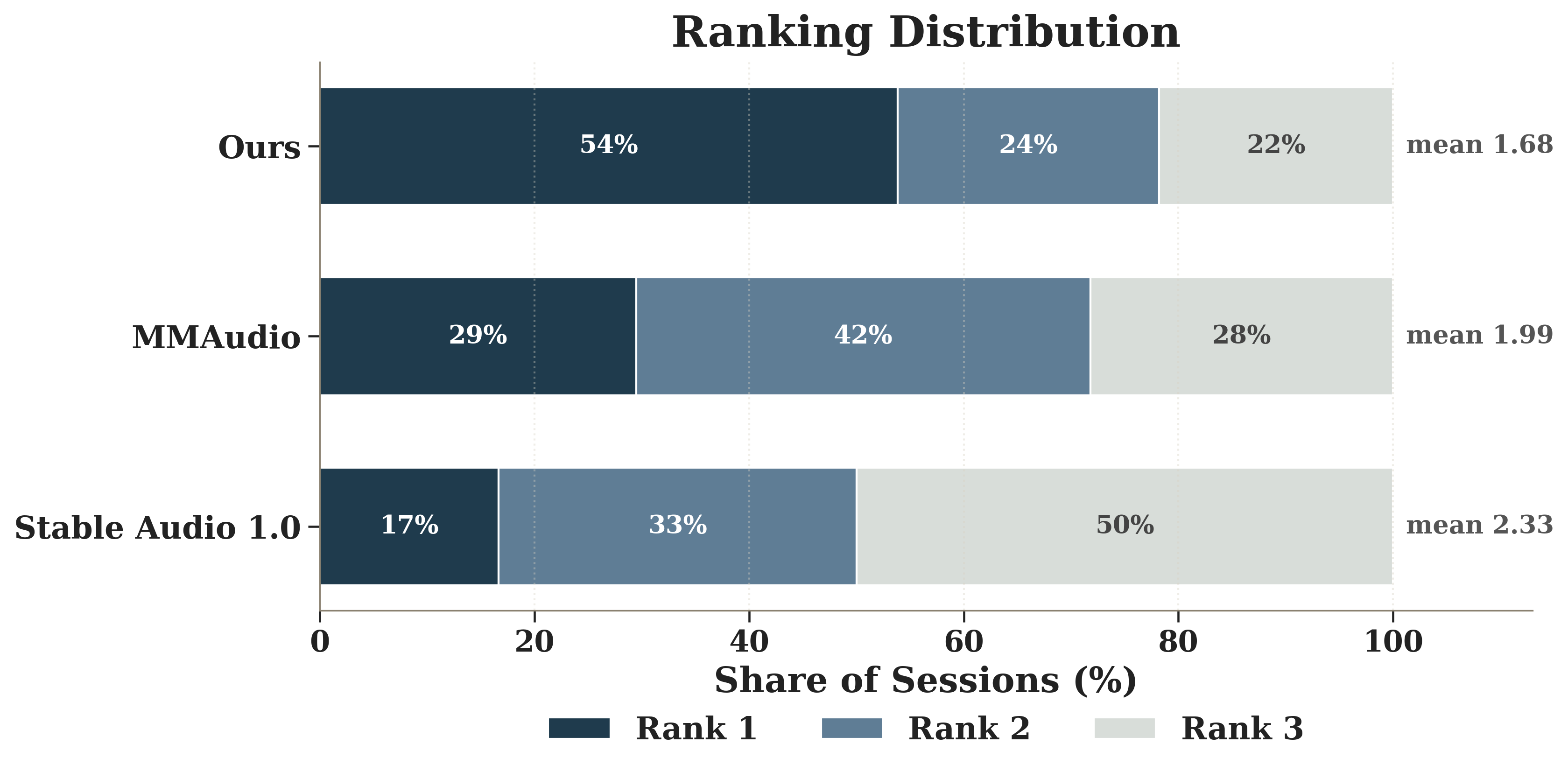}
        \captionof{figure}{Original long-horizon VLM rank distribution for three methods.}
        \label{fig:rank_vlm}
    \end{minipage}

\end{figure*}
\subsection{Duration-matched automated comparison}
\label{sec:short_vlm_app}
The long-horizon studies evaluate audio--visual consistency and
listener-motion-dependent behavior over 20--40\,s trajectories. The
released See-2-Sound and OmniAudio configurations used in our comparisons
provide approximately 5--10\,s outputs. We therefore evaluate these
baselines in a separate automated comparison in which every candidate is
uniformly truncated to 5\,s (Table~\ref{tab:vlm_eval_5s}).

The comparison reports model-assigned mean MOS, mean rank, Rank-1 rate,
Top-2 rate, and Borda score for all five methods. It does not add human
participants or extend the original three-method human study.
The original long-horizon VLM results are retained in
Table~\ref{tab:vlm_eval_summary} and
Figures~\ref{fig:per_cat_mos_vlm} and~\ref{fig:rank_vlm}; those plots do not
represent the new five-method comparison. Ranking and Borda values from
the three- and five-candidate studies should not be directly compared
because both the candidate set and evaluated duration differ.

\section{Metric Definition}

\subsection{Semantic Metrics}
\label{sec:semantic_metric_app}
We use two complementary semantic evaluation protocols. The first is an embedding-based protocol, which scores text--audio and image--audio agreement using LAION-CLAP and ImageBind. The second is a caption-based protocol, which uses Audio Flamingo 2 to describe the generated audio and then measures semantic agreement between the generated caption and the scene prompt.

\paragraph{Session inputs and prompt resolution.}
For each session, for the embedding-based protocol, the scene text is used directly as the text query unless optional LLM rewrite is explicitly enabled. For the caption-based protocol, the default reference text is an audio-relevant filtered version of the scene prompt, in which a fixed list of mostly visual words (e.g., colors, lighting, view/composition terms, and rendering descriptors) is removed before similarity scoring.

\subsubsection{LAION-CLAP.}
For the CLAP-based runs reported in this work, we use the official LAION-CLAP implementation with the checkpoint \texttt{music\_speech\_audioset\_epoch\_15\_esc\_89.98.pt}. Given a text query $x$ and an audio waveform $a$, the evaluator extracts a normalized text embedding $\mathbf{t}$ and a normalized audio embedding $\mathbf{u}$ and reports their cosine similarity
\[
s_{\mathrm{CLAP}}(x,a) = \cos(\mathbf{t}, \mathbf{u}).
\]
At the session level, audio is clipped to the first $20$\,s before scoring in order to keep evaluation length consistent across methods. The resulting scalar is stored as the session's \texttt{text\_audio} score, and dataset-level results are obtained by averaging over all successfully processed sessions. 

\subsubsection{ImageBind text--audio and image--audio scores.}
We complement CLAP with ImageBind using the released model. The text--audio score is computed analogously to CLAP: the scene text and the session audio are embedded, $\ell_2$-normalized, and compared with cosine similarity,
\[
s_{\mathrm{IB,text}}(x,a) = \cos(\mathbf{z}_{x}, \mathbf{z}_{a}).
\]
For image--audio evaluation, we convert the session panorama from equirectangular projection into multiple perspective views before embedding. In the default session-level configuration used in our large runs, we use the four horizontal views \emph{front}, \emph{right}, \emph{back}, and \emph{left}; an optional flag also enables \emph{up} and \emph{down} views. Each perspective image is rendered at $512 \times 512$, then resized and center-cropped to the CLIP/ImageBind input resolution of $224 \times 224$. If $\{\mathbf{z}_{v_i}\}_{i=1}^{N}$ are the normalized image-view embeddings and $\mathbf{z}_{a}$ is the normalized audio embedding, we report
\[
s_{\mathrm{IB,img}}^{\mathrm{mean}} = \frac{1}{N}\sum_{i=1}^{N} \cos(\mathbf{z}_{v_i}, \mathbf{z}_{a}),
\]

\subsubsection{Audio Flamingo 2 caption-based evaluation.}
Our caption-based semantic metric uses the released Audio Flamingo 2 checkpoint \texttt{nvidia/audio-flamingo-2-0.5B}. Audio is converted to mono at $16$\,kHz, normalized to $[-1,1]$, divided into CLAP-style windows according to the AF2 inference configuration, and zero-padded when needed. Caption generation is conditioned on the text prompt \texttt{"Caption the input audio."}. The generated text is then cleaned by removing separator, padding, and end-of-chunk tokens.

\paragraph{Prompt--caption similarity.}
After AF2 produces a caption $\hat{c}$ for a session, we compare it against the audio-relevant reference prompt $\tilde{x}$ using a SentenceTransformer encoder. In our runs, this encoder is \texttt{all-MiniLM-L6-v2}. Let $\phi(\cdot)$ denote the normalized sentence embedding. The final prompt--caption score is
\[
s_{\mathrm{AF2}}(\tilde{x}, \hat{c}) = \cos\!\bigl(\phi(\tilde{x}), \phi(\hat{c})\bigr).
\]
At the dataset level, we report the mean of these per-session similarities over all successfully processed sessions.

\subsubsection{Reasons for the three metrics}
The three metrics emphasize different aspects of semantic fidelity. LAION-CLAP and ImageBind text--audio are direct cross-modal retrieval scores against the original scene description; ImageBind image--audio further tests whether the generated sound is consistent with the scene appearance under multiple panorama views; and Audio Flamingo 2 measures whether a large audio-captioning model can recover a textual description that remains semantically close to the intended prompt. We therefore treat these metrics as complementary rather than interchangeable.

\subsubsection{Scene-level uncertainty}
\label{sec:semantic_ci_app}
We report bootstrap 95\% confidence intervals for our semantic scores over
the original 80 generated scenes (Table~\ref{tab:semantic_ci}). The means
are shown at the precision reported in the uncertainty analysis. These
intervals characterize the scene-level evaluation; they do not constitute
a paired significance test against the baselines.

\begin{table}[t]
\centering
\small
\setlength{\tabcolsep}{10pt}
\caption{Semantic scores with bootstrap 95\% confidence intervals over
the original 80 scenes.}
\label{tab:semantic_ci}
\begin{tabular}{@{}lrc@{}}
\toprule
Metric & Mean & 95\% CI \\
\midrule
CLAP & 0.317 & $[0.286,0.348]$ \\
IB-Text & 0.220 & $[0.130,0.260]$ \\
IB-Image & 0.262 & $[0.142,0.302]$ \\
Caption & 0.276 & $[0.230,0.322]$ \\
\bottomrule
\end{tabular}
\end{table}

\subsection{Spatial Metrics}
\label{sec:spatial_metric_app}
 \subsubsection{Windowed ILD}
Since binaural cues are usually treated as locally stationary, we use the standard windowed ILD metrics for evaluation.
For a binaural signal with left and right channels \(y_L\) and \(y_R\), we compute ILD in short windows indexed by \(m\):
\begin{equation}
    \mathrm{ILD}(m) = 10\log_{10}
    \frac{\sum_{t\in \Omega_m} y_L(t)^2 + \varepsilon}
         {\sum_{t\in \Omega_m} y_R(t)^2 + \varepsilon},
\end{equation}
where \(\Omega_m\) is the analysis window and \(\varepsilon\) prevents numerical instability. Here we use 0.2s for window size and ensuring 50\% overlapping between windows.

\subsubsection{ILD proxy fit.}
Prior acoustic studies show that ILD varies systematically with source azimuth and can often be approximated by a sinusoidal trend in the horizontal plane; its magnitude also increases as the source becomes closer, especially for lateral directions~\cite{kim2010acoustic}. Motivated by this, we define a simple geometry-based proxy. We fit the geometry-based diagnostic proxy,
\begin{equation}
    \label{eq:ild_proxy}
    \mathrm{ILD}_{\mathrm{proxy}}(t)
    = k\frac{\sin(\theta(t))}{d(t)^\alpha}+b,
\end{equation}
where \(\theta(t)\) is the signed horizontal listener-source angle and \(d(t)\) is listener-source distance.
In the implementation, $\alpha$ is selected by grid search over
\[
\alpha \in \{0, 0.05, 0.10, \dots, 2.50\},
\]
and for each fixed $\alpha$, the parameters \(k\) and  \(b\) are obtained by least squares. We report three metrics: ILD Corr., the Pearson correlation between the measured ILD curve and the fitted proxy; ILD $R^2$, the fraction of ILD variance explained by the proxy; and mean $\mathrm{ILD}_{\mathrm{abs}}$, the average absolute ILD magnitude in dB, where larger values indicate stronger left-right lateralization. Scores are averaged across sources and scenes, with results summarized in Table~\ref{tab:spatial_eval}. For mono-channel baselines such as MMAudio, spatial channel cues are absent, so these metrics are not applicable.

\begin{table}[t]
\centering
\small
\setlength{\tabcolsep}{7pt}
\caption{Original ILD diagnostics. Correlation and $R^2$ measure agreement
with the fitted geometry-based proxy. Mean absolute ILD (dB) measures
lateralization strength, not directional accuracy by itself.}
\label{tab:spatial_eval}
\begin{tabular}{@{}lrrr@{}}
\toprule
Method & ILD Corr. & ILD $R^2$ & Mean $\mathrm{ILD}_{\mathrm{abs}}$ \\
\midrule
See-2-Sound & 0.1763 & 0.0408 & 1.6938 \\
OmniAudio & 0.3050 & 0.1270 & 2.9048 \\
Stable Audio 1.0 & 0.1393 & 0.0291 & 1.2081 \\
MMAudio & -- & -- & -- \\
Ours & 0.8958 & 0.8097 & 9.4735 \\
\bottomrule
\end{tabular}
\end{table}

\subsubsection{Energy-ranking consistency.}
For each scene, let \(E_i\) be the rendered energy of source \(i\), and let \(\bar{d}_i\) be its average distance to the listener along the trajectory. Pairwise rank accuracy is the fraction of source pairs whose predicted acoustic-energy ordering agrees with the inverse-distance ordering:
\begin{equation}
    \mathrm{Acc}_{\mathrm{pair}} =
    \frac{1}{|\mathcal{P}|}\sum_{(i,j)\in\mathcal{P}}
    \mathbf{1}\left[(E_i-E_j)(\bar{d}_j-\bar{d}_i)>0\right].
\end{equation}
We also report Spearman rank correlation between \(E_i\) and \(-\bar{d}_i\). These metrics are computed only when per-source renders are available and used for our pipeline in the ablation study in Section~\ref{sec:ablation}.

\begin{figure*}
    \centering
    \includegraphics[width=0.99\linewidth]{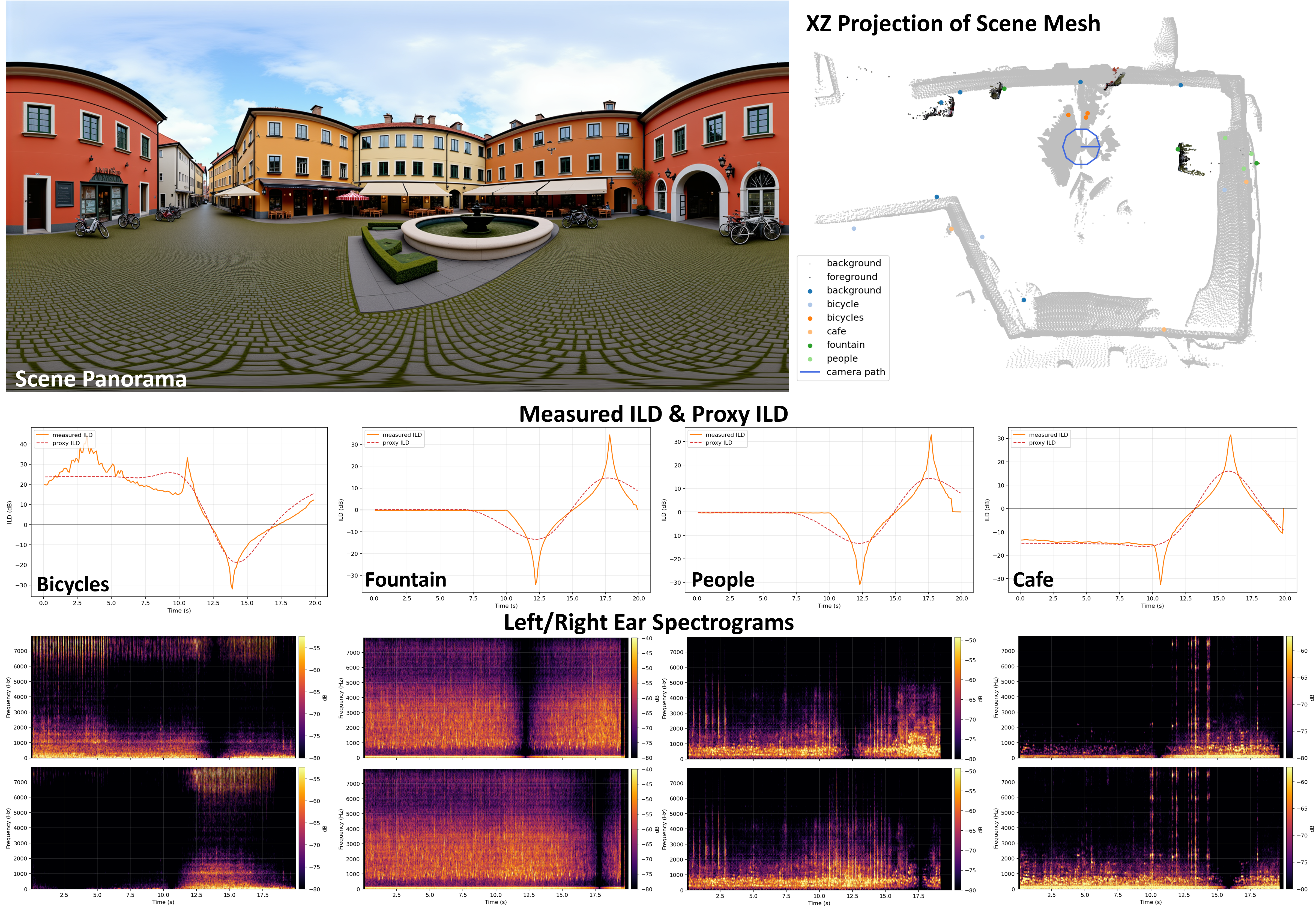}
    \caption{
    Qualitative spatial analysis for one generated scene. We visualize the panorama, the top-down source layout, the proxy ILD curve induced by listener--source geometry, the measured ILD from the rendered binaural audio, and channel-wise spectrograms. The measured ILD follows the geometric trend, indicating that the soundtrack changes consistently with listener motion.
    }
    \label{fig:ild_qual_1}
\end{figure*}
\subsection{Ground-truth-referenced direction of arrival}
\label{sec:doa_app}

\paragraph{Reference azimuths.}
Ground truth is defined relative to the constructed audible world.
Each evaluated localized semantic label is one logical source whose
position $\mathbf{s}_i$ is the centroid of its associated emitters.
Background emitters are excluded from the reference source set in this
evaluation. Listener position and orientation are interpolated from the
known trajectory. Let $\mathbf{l}(t)$, $\mathbf{f}(t)$, and $\mathbf{u}(t)$
be listener position, forward direction, and up direction. With
\begin{equation}
\mathbf{r}(t)=
\frac{\mathbf{f}(t)\times\mathbf{u}(t)}
     {\|\mathbf{f}(t)\times\mathbf{u}(t)\|},
\end{equation}
the reference horizontal azimuth is
\begin{equation}
\theta_i(t)=\operatorname{wrap}_{[-180,180)}\!\left[
\frac{180}{\pi}\operatorname{atan2}\!\left(
(\mathbf{s}_i-\mathbf{l}(t))\cdot\mathbf{r}(t),
(\mathbf{s}_i-\mathbf{l}(t))\cdot\mathbf{f}(t)
\right)\right].
\label{eq:doa_reference}
\end{equation}
Positive azimuth is listener-right. This reference tests directional
consistency with assigned source positions; it is not an independently
measured real-world source-position or depth ground truth.

\paragraph{Isolated-source estimator.}
We estimate full-circle horizontal DoA from binaural windows of at least
0.5\,s using multi-band ILD, GCC-PHAT interaural time difference, a
low-weight pinna-spectral cue for front--back disambiguation, and circular
trajectory decoding. Calibration uses 36 azimuths at $10^\circ$ intervals
over the full circle. Table~\ref{tab:isolated_doa_ci} reports the resulting
angular errors and accuracy within $15^\circ$.

\begin{table}[t]
\centering
\small
\setlength{\tabcolsep}{9pt}
\caption{Full-circle isolated-source DoA recovery for our method.
Angular errors are in degrees; accuracy is a percentage.}
\label{tab:isolated_doa_ci}
\begin{tabular}{@{}lrc@{}}
\toprule
Metric & Estimate & 95\% CI \\
\midrule
MAE $\downarrow$ & $6.91^\circ$ & $[3.51^\circ,12.04^\circ]$ \\
Median error $\downarrow$ & $2.96^\circ$ & $[2.64^\circ,3.31^\circ]$ \\
Accuracy@$15^\circ$ $\uparrow$ & 96.1\% & $[92.8\%,98.9\%]$ \\
\bottomrule
\end{tabular}
\end{table}

\paragraph{Mixture activity mask.}
Directions are estimated from the final mixed soundtracks. A reference
source is active only when its dry signal is above $-60$\,dBFS and within
40\,dB of its session-level maximum. The same reference activity mask is
used for every method, avoiding method-dependent selection of favorable
windows. Exclusion from the reference inventory should not be interpreted
as source separation of the mixed waveform.

\paragraph{Channel-specific mixture estimators.}
Our binaural outputs use held-out GSound calibration and circular ILD
peak detection. See-2-Sound~\cite{dagli2024see2sound} uses its $\pm x$ and
$\pm z$ microphones to estimate horizontal direction. For
OmniAudio~\cite{liu2025omniaudiogeneratingspatialaudio}, we use first-order
Ambisonic active intensity and convert its coordinates to the same
positive-listener-right convention. These channel-specific estimators
respect the different output geometries; the comparison therefore
measures recoverable direction under each method's estimator.
MMAudio is evaluated in its released mono configuration without post-hoc
spatialization. Stable Audio 1.0 produces generic stereo without a defined
binaural HRTF, array geometry, or listener-centric coordinate reference.
Neither is assigned an absolute-azimuth DoA score.

\paragraph{Matching and metrics.}
We perform one-to-one Hungarian assignment between predicted and active
reference directions using circular angular distance,
\begin{equation}
 d_{\mathrm{circ}}(\hat\theta,\theta)
 = \left|\operatorname{wrap}_{[-180,180)}(\hat\theta-\theta)\right|.
\end{equation}
Matched MAE measures error over assigned direction pairs. At angular
tolerance $\delta\in\{15^\circ,45^\circ\}$, a pair within tolerance counts
as a true positive, while remaining predictions and references contribute
false positives and false negatives. For the corresponding counts,
\begin{equation}
 F_1@\delta =
 \frac{2\,\mathrm{TP}_{\delta}}
      {2\,\mathrm{TP}_{\delta}+\mathrm{FP}_{\delta}+\mathrm{FN}_{\delta}}.
\end{equation}
The results in Table~\ref{tab:mixture_doa} show lower matched error and
higher $F_1@15^\circ$ for our method, but higher $F_1@45^\circ$ for
See-2-Sound. Matched MAE must be interpreted alongside $F_1$, since a low
error over assigned pairs alone does not establish recovery of all sources.
Mixture DoA is complementary to, rather than a substitute for, perceptual
source-distinctness evaluation.

\section{Runtime Breakdown}
\label{sec:time_breakdown}
\begin{figure}[t]
    \centering
    \includegraphics[width=0.72\linewidth]{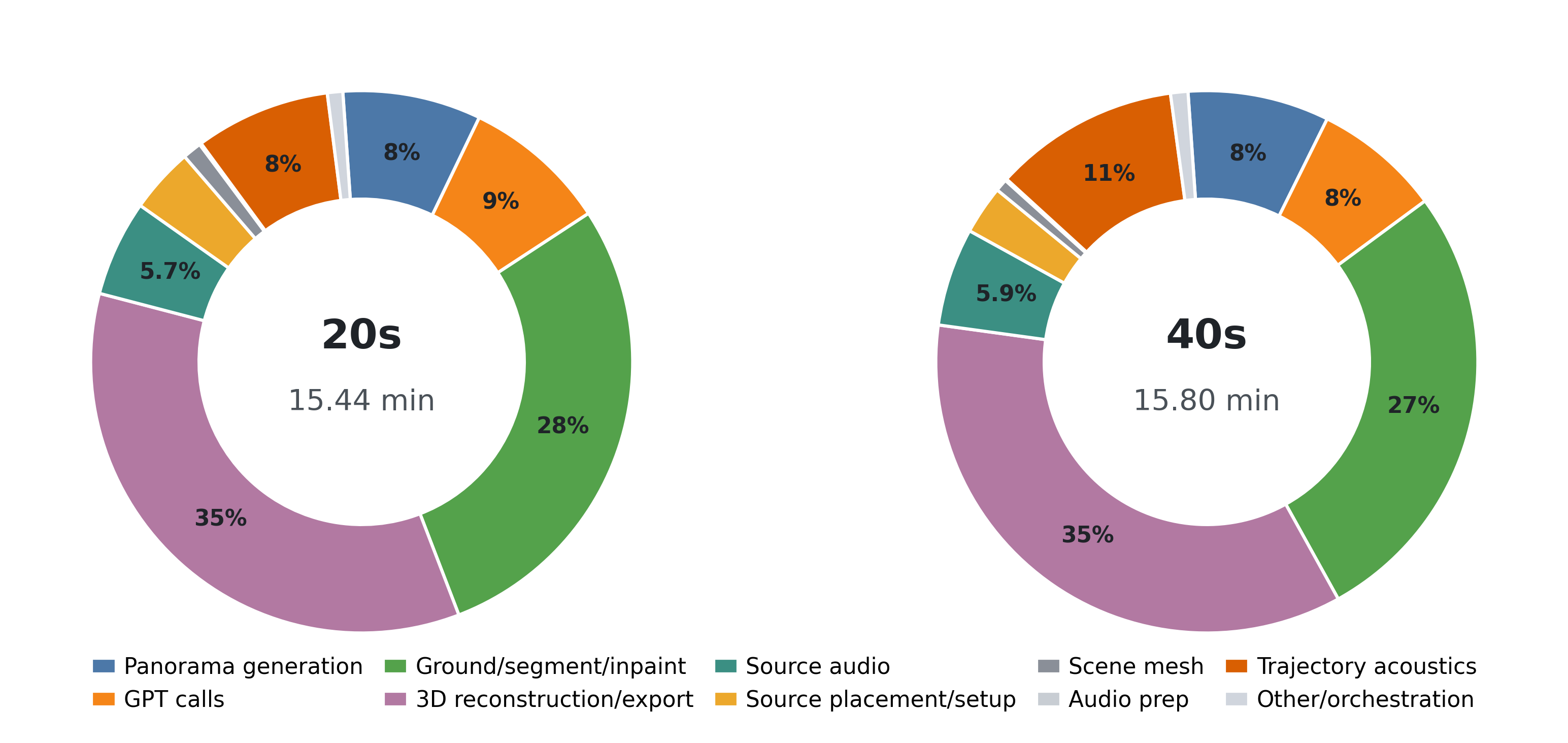}
    \caption{Runtime distribution of the full generation pipeline for 20s and 40s trajectories.}
    \label{fig:timing}
\end{figure}

We measure the runtime of the full pipeline on a single GPU with a network-mounted hard drive. The reported timing includes sound generation, acoustic simulation, GPT network calls, and HunyuanWorld generation. The main bottleneck is 3D world generation and file saving, while our agentic sound generation and acoustic simulation account for a relatively small fraction of the total runtime. The results are provided in Figure~\ref{fig:timing} and Table~\ref{tab:full_pipeline_runtime}.
\begin{table}[t]
\centering
\setlength{\tabcolsep}{4pt}
\renewcommand{\arraystretch}{1.05}
\caption{Full runtime breakdown of our generation pipeline. The dominant costs are perception and 3D world reconstruction/export, while source-audio generation and acoustic rendering account for a smaller fraction of the total runtime.}
\label{tab:full_pipeline_runtime}
\begin{tabular}{lrrrr}
\toprule
\textbf{Operation} &
\multicolumn{2}{c}{\textbf{20s}} &
\multicolumn{2}{c}{\textbf{40s}} \\
\cmidrule(lr){2-3}\cmidrule(lr){4-5}
& \textbf{Sec.} & \textbf{Share} & \textbf{Sec.} & \textbf{Share} \\
\midrule
Panorama generation      & 76.50  & 8.3\%  & 80.00  & 8.4\% \\
GPT calls                & 80.04  & 8.6\%  & 71.74  & 7.6\% \\
Ground/segment/inpaint   & 263.00 & 28.4\% & 256.50 & 27.1\% \\
3D reconstruction/export & 323.41 & 34.9\% & 334.06 & 35.2\% \\
Source audio             & 53.00  & 5.7\%  & 55.50  & 5.9\% \\
Source placement/setup   & 36.00  & 3.9\%  & 27.50  & 2.9\% \\
Scene mesh               & 10.50  & 1.1\%  & 7.00   & 0.7\% \\
Audio prep               & 1.00   & 0.1\%  & 1.00   & 0.1\% \\
Trajectory acoustics     & 74.70  & 8.1\%  & 105.07 & 11.1\% \\
Other/orchestration      & 8.52   & 0.9\%  & 9.83   & 1.0\% \\
\midrule
\textbf{Total}           & \textbf{926.67} & \textbf{100.0\%} 
                         & \textbf{948.20} & \textbf{100.0\%} \\
\bottomrule
\end{tabular}
\end{table}

\subsection{Controlled scaling with a cached world}
\label{sec:cached_scaling_app}
We hold the reconstructed world fixed and vary trajectory duration or the
rendering-source cap. Panorama generation, segmentation, inpainting, and
reconstruction are therefore unchanged; source setup and acoustic
propagation constitute the varying costs. These measurements are distinct
from the end-to-end timings in Table~\ref{tab:full_pipeline_runtime}.

\begin{table}[t]
\centering
\small
\caption{Controlled cached-world scaling. Left: 32 sources with varying
trajectory duration. Right: a fixed 20\,s trajectory with varying source
cap. Times are in seconds; the $\pm$ terms reproduce the timing
variability reported with the experiments. Source caps concern rendering
emitters, not semantic labels.}
\label{tab:cached_world_scaling}
\begin{minipage}[t]{0.47\linewidth}
\centering
\begin{tabular}{@{}rr@{}}
\toprule
Duration (s) & Reported time (s) \\
\midrule
5 & $21.36 \pm 9.86$ \\
10 & $34.42 \pm 10.61$ \\
20 & $46.34 \pm 13.45$ \\
40 & $57.67 \pm 19.39$ \\
80 & $106.20 \pm 31.77$ \\
160 & $184.14 \pm 18.08$ \\
\bottomrule
\end{tabular}
\end{minipage}\hfill
\begin{minipage}[t]{0.47\linewidth}
\centering
\begin{tabular}{@{}rr@{}}
\toprule
Source cap & Reported time (s) \\
\midrule
1 & $22.32 \pm 11.06$ \\
2 & $31.09 \pm 10.42$ \\
4 & $32.09 \pm 9.43$ \\
8 & $34.49 \pm 10.92$ \\
16 & $40.00 \pm 10.06$ \\
32 & $46.34 \pm 13.45$ \\
64 & $63.17 \pm 17.75$ \\
128 & $92.10 \pm 17.85$ \\
\bottomrule
\end{tabular}
\end{minipage}
\end{table}

Table~\ref{tab:cached_world_scaling} shows increasing runtime with longer
trajectories and larger source caps over the tested range, while the
reconstructed world can be reused. These measurements do not establish
an asymptotic complexity bound or real-time performance. GSound is
CPU-based, so absolute runtime and variability also depend on available
threads and concurrent CPU load.

\subsection{Controlled moving-source rendering}
\label{sec:moving_sources_app}
To test the rendering interface beyond fixed source positions, we provide
heuristic trajectories for the sources and the listener directly to
GSound. Source audio is generated by our pipeline, and each source starts
at the position used in the original scene. Both listener and source
trajectories last 20\,s. Geometry remains static: the experiment does not
construct moving visual objects, infer source motion, or reconstruct
dynamic geometry.

\begin{table}[t]
\centering
\small
\setlength{\tabcolsep}{7pt}
\caption{Moving-source rendering with supplied 20\,s source and listener
trajectories in a static reconstructed world. Warm-render times and
paired slowdowns are reported as measured; memory is in MiB. This is a
rendering-stage proof of capability, not end-to-end dynamic-world
construction.}
\label{tab:moving_sources}
\begin{tabular}{@{}rrrr@{}}
\toprule
Moving sources & Warm render (s) & Paired slowdown & Memory (MiB) \\
\midrule
1 & 24.51 & $1.00\times$ & 544.5 \\
2 & 51.81 & $2.12\times$ & 558.2 \\
4 & 65.98 & $2.66\times$ & 586.4 \\
8 & 77.86 & $3.22\times$ & 641.4 \\
16 & 89.95 & $3.67\times$ & 753.6 \\
32 & 106.09 & $4.33\times$ & 978.6 \\
\bottomrule
\end{tabular}
\end{table}

The test demonstrates that generated source assets can be rendered along
provided trajectories, reaching 32 moving sources with a warm-render time
of 106.09\,s and reported memory of 978.6\,MiB. Automatic dynamic-scene
construction remains future work. The timings do not by themselves
validate motion estimation, visual synchronization, or Doppler behavior.

\section{User Study Forms}
Here we show the user study google form setup we used. 
\begin{figure}[h]
    \centering
    \includegraphics[width=0.5\linewidth]{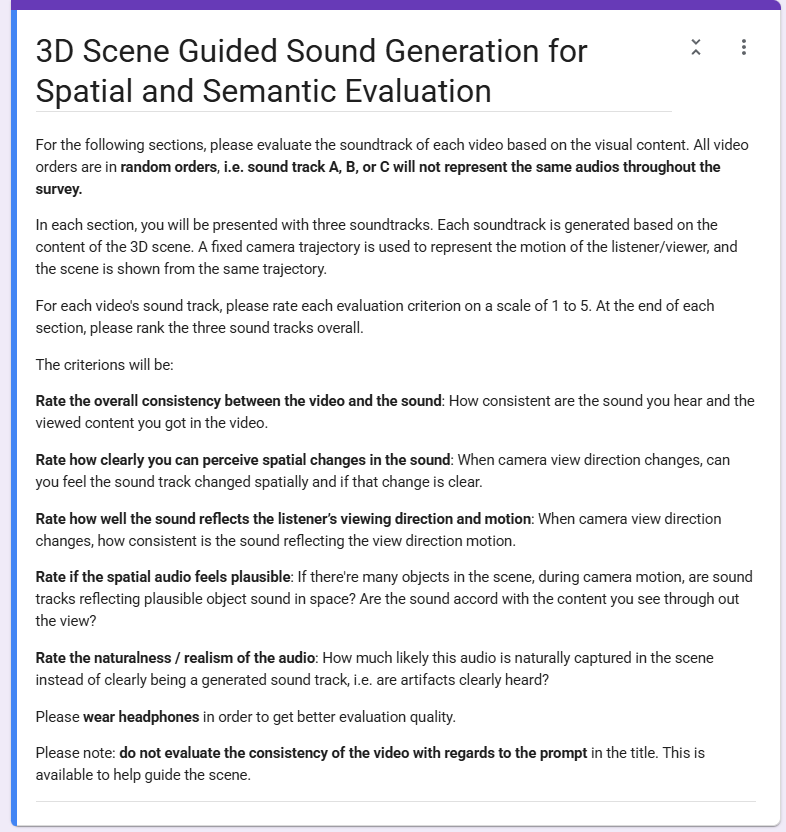}
    \caption{User Instruction Page}
    \label{fig:user1}
\end{figure}

\begin{figure*}
    \centering
    \includegraphics[width=0.99\linewidth]{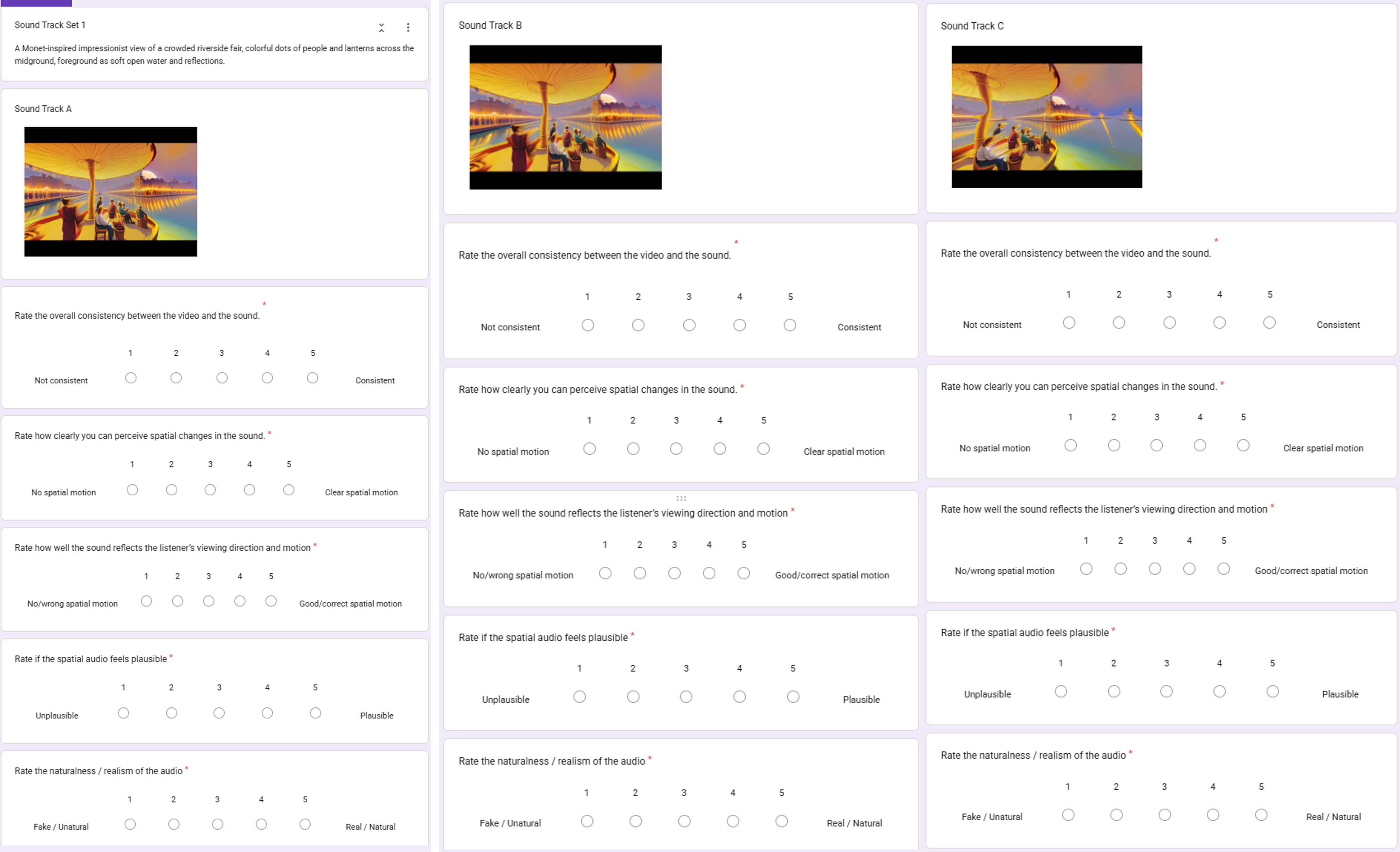}
    \caption{User Study MOS Questions}
    \label{fig:user2}
\end{figure*}

\begin{figure*}
    \centering
    \includegraphics[width=0.7\linewidth]{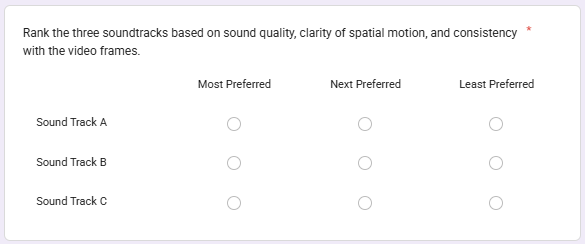}
    \caption{User Study Ranking Question}
    \label{fig:user3}
\end{figure*}

\end{document}